%% file: main.tex
\documentclass[conference]{IEEEtran}
\IEEEoverridecommandlockouts
\usepackage{cite}
\usepackage{amsmath,amssymb,amsfonts}
\usepackage{algorithmic}
\usepackage{graphicx}
\usepackage{hyperref}
\usepackage{textcomp}
\usepackage{xcolor}
\usepackage{subcaption}
\usepackage{booktabs}
\usepackage{fancyhdr}

\newcommand{\etal}{\textit{et~al}.}
\newcommand{\ie}{\textit{i}.\textit{e}., }

\usepackage{xcolor}

\DeclareMathOperator{\sg}{sg}

\def\BibTeX{{\rm B\kern-.05em{\sc i\kern-.025em b}\kern-.08em
    T\kern-.1667em\lower.7ex\hbox{E}\kern-.125emX}}

\fancypagestyle{first_page_style}
{
   \fancyhf{}
   \lfoot{\color{gray}\scriptsize © 2025 IEEE. Personal use of this material is permitted. Permission from IEEE must be obtained for all other uses, in any current or future media, including reprinting/republishing this material for advertising or promotional purposes, creating new collective works, for resale or redistribution to servers or lists, or reuse of any copyrighted component of this work in other works.}
}

\begin{document}

\title{Denoising Diffusion Probabilistic Model for\\Point Cloud Compression at Low Bit-Rates}

\author{\IEEEauthorblockN{
Gabriele Spadaro$^{1,2}$,
Alberto Presta$^{2}$,
Jhony H. Giraldo$^1$, 
Marco Grangetto$^2$,
Wei Hu$^3$,\\
Giuseppe Valenzise$^{4}$,
Attilio Fiandrotti$^{1,2}$,
Enzo Tartaglione$^1$
}
\IEEEauthorblockA{$^1$LTCI, T\'el\'ecom Paris, Institut Polytechnique de Paris, France~~
$^2$University of Turin, Italy}
\IEEEauthorblockA{$^3$Wangxuan Institute of Computer Technology,
Peking University}
\IEEEauthorblockA{$^4$Université Paris-Saclay, CNRS, CentraleSupélec, Laboratoire des Signaux et Systèmes}
}

\maketitle

\begin{abstract}
Efficient compression of low-bit-rate point clouds is critical for bandwidth-constrained applications. However, existing techniques mainly focus on high-fidelity reconstruction, requiring many bits for compression.
This paper proposes a ``Denoising Diffusion Probabilistic Model'' (DDPM) architecture for point cloud compression (DDPM-PCC) at low bit-rates. 
A PointNet encoder produces the condition vector for the generation, which is then quantized via a learnable vector quantizer. This configuration allows to achieve a low bitrates while preserving quality. Experiments on ShapeNet and ModelNet40 show improved rate-distortion at low rates compared to standardized and state-of-the-art approaches. We publicly released the code at \href{https://github.com/EIDOSLAB/DDPM-PCC}{https://github.com/EIDOSLAB/DDPM-PCC}.
\end{abstract}
\fancyhf{}
\renewcommand{\headrulewidth}{0pt}
\cfoot{\thepage}
\thispagestyle{first_page_style}

\input{sec/1_intro}

\input{sec/2_related}
\input{sec/3_method}

\input{sec/4_experiments}

\input{sec/5_conclusion}

\bibliographystyle{IEEEbib}
\bibliography{icme2025references}

\end{document}

%% file: sec/1_intro.tex
\section{Introduction}
\label{sec:intro}

Recent advances in 3D imaging have driven interest in Point Clouds Compression (PCC) technologies. 
Namely, the complexity of storing and delivering point cloud 
geometry poses significant challenges due to their spatial complexity.
For this reason, traditional codecs based on effective data structure have been proposed.  
The MPEG Geometry-based Point Cloud Compression (G-PCC)~\cite{gpcc} standard, for instance, aims at achieving high-quality compression of static, highly detailed point clouds, leveraging grid-based and spatial subdivision techniques.
Similarly, Google released DRACO~\cite{draco}, an open-source library for point cloud compression.

The recent success of neural models in image and video compression~\cite{balle17,liu2023learned} prompted learning-based approaches for PCC.
In~\cite{quach2019learning}, the authors introduced a pair of specific analysis-synthesis convolutional transforms for PCC compression.
In~\cite{yan2019deep} the authors proposed a PointNet-based~\cite{pointnet} encoder, while~\cite{shen2024multi} adopted a PointNet++~\cite{pointnet++} module. More recently, attention-based point cloud deconvolution modules have been proposed to improve reconstruction by addressing localized artifacts from linear interpolation~\cite{shen2024multi}. 
Li~\etal~\cite{cot_pcc}, instead, introduced an optimal transport system to learn the distribution transformation from original to reconstructed point clouds, 
while He~\etal~\cite{dpcc} highlighted that maintaining local density is crucial for achieving faithful reconstruction, proposing 
a novel deep learning-based autoencoder designed to maintain this local property. 
Moreover, hybrid solutions were also proposed to synergistically employ deep learning and octree-based representations~\cite{que2021voxelcontext, fu2022octattention, morell2014geometric}, 
or following a voxel-based approach~\cite{quach2019learning, quach2020improved,wang2021lossy, wang2021multiscale}, where a point cloud is represented as a structured grid of volumetric cells (voxels).
All the above methods target high fidelity reconstruction, struggling to preserve quality at bandwidths practically available on real-world channels (as visualized in Fig.~\ref{fig:teaser}), motivating the present work.

\input{plots/teaser}

When working at lower bit-rates, a recent yet effective approach to perform compression is towards generative models like diffusion models~\cite{ddpm}. 
Seminal works like~\cite{yang2023lossy} 
exploit a conditional diffusion model for image compression. Here, the reverse diffusion process is conditioned using a latent representation extracted via analysis transform. 
In~\cite{careil2023towards}, a stable diffusion model was conditioned on text embeddings extracted from a pre-trained image captioning model and transmitted to the decoder. Here, a pre-trained latent diffusion model (LDM) was used to encode the image into a latent representation and then quantized via a hyperprior autoencoder. 
Although effective, these approaches are designed for images with a fixed spatial structure, which is fundamentally
different from the irregular and unstructured nature of point
clouds. To the best of our knowledge, no attempts have been
made to employ such methods specifically for point cloud
data.

In this work, we propose a Denoising Diffusion Probabilistic Model architecture for PCC (DDPM-PCC). Specifically, we use PointNet~\cite{pointnet} to map the initial point cloud to a latent representation for initial compression, followed by vector quantization. The resulting quantized representation is then passed to the decoder, which conditions a diffusion process to produce the final reconstruction.

Our contributions can be summarized as follows:
\begin{itemize}
    \item To the best of our knowledge, we are the first to employ a diffusion-based generative model for geometry point cloud compression, detailed in Sec.~\ref{subsec:comp_model}.
    \item We adopted a compression solution based on a learnable vector quantizer to compress the latent space (Sec.~\ref{subsec:econde_latent}). 
    \item Our experimental results, presented in Sec.~\ref{subsec:rd}, show that our method improves the encoding efficiency at a low bit rate with respect to other state-of-the-art approaches.
\end{itemize}

%% file: plots/teaser.tex
\begin{figure}[!t]
    \centering
    \begin{subfigure}{0.49\columnwidth}
        \centering
        \includegraphics[width=.8\linewidth]{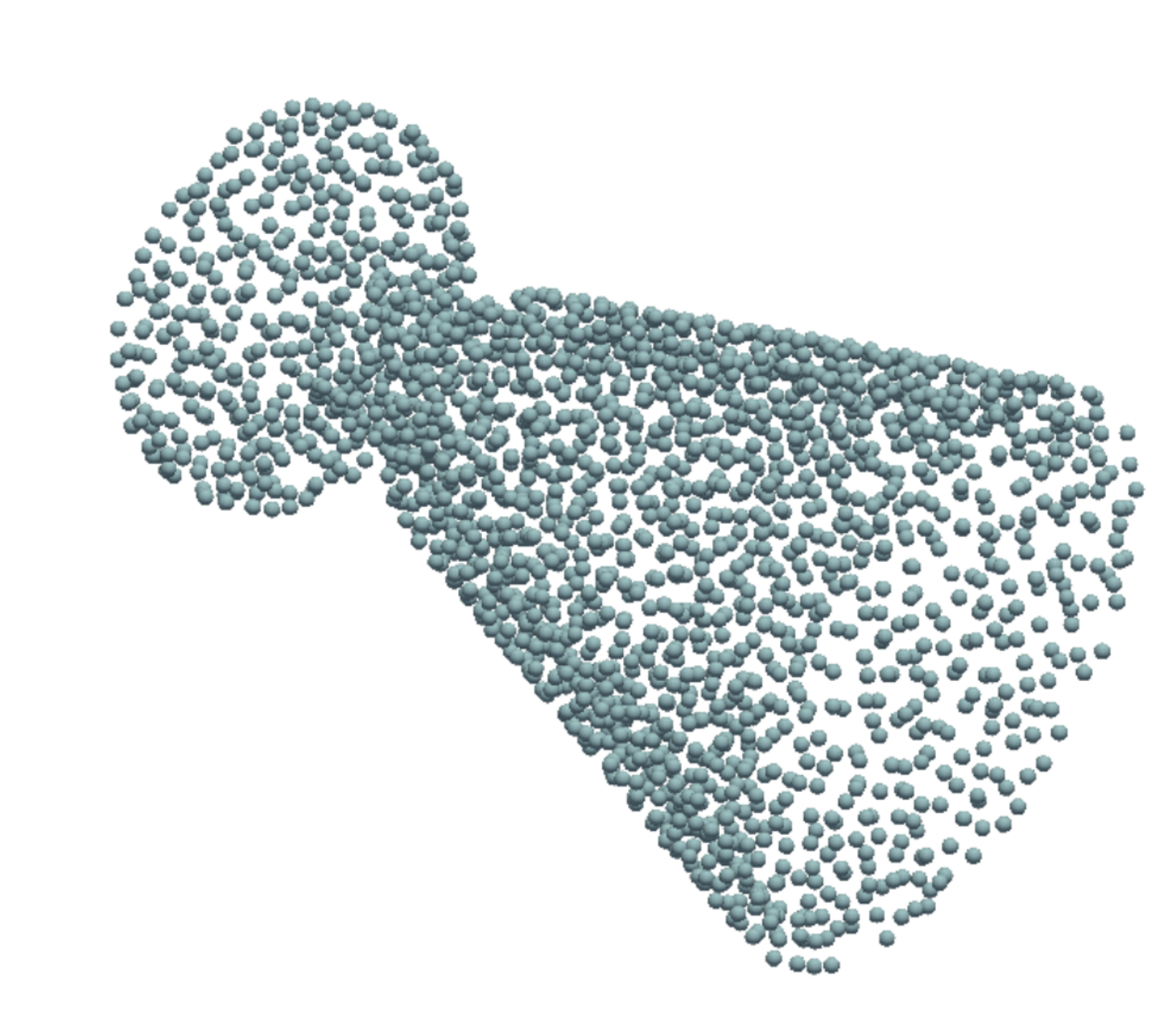} 
        \caption*{Ground Truth}
    \end{subfigure}
    \hfill
    \begin{subfigure}{0.49\columnwidth}
        \centering
        \includegraphics[width=.8\linewidth]{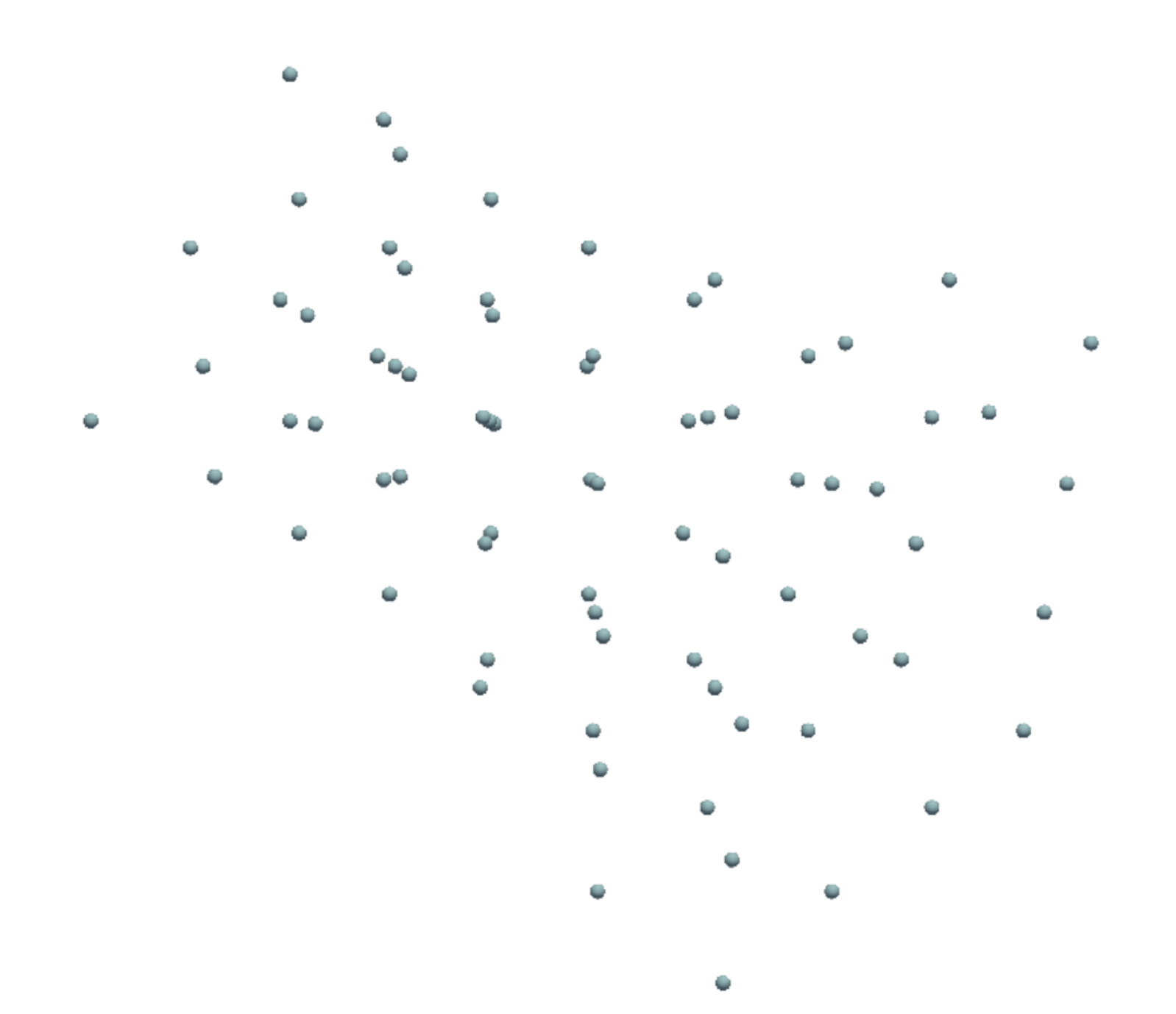} 
        \caption*{G-PCC (bpp: 0.10)}
    \end{subfigure}
    \begin{subfigure}{0.49\columnwidth}
        \centering
        \includegraphics[width=.8\linewidth]{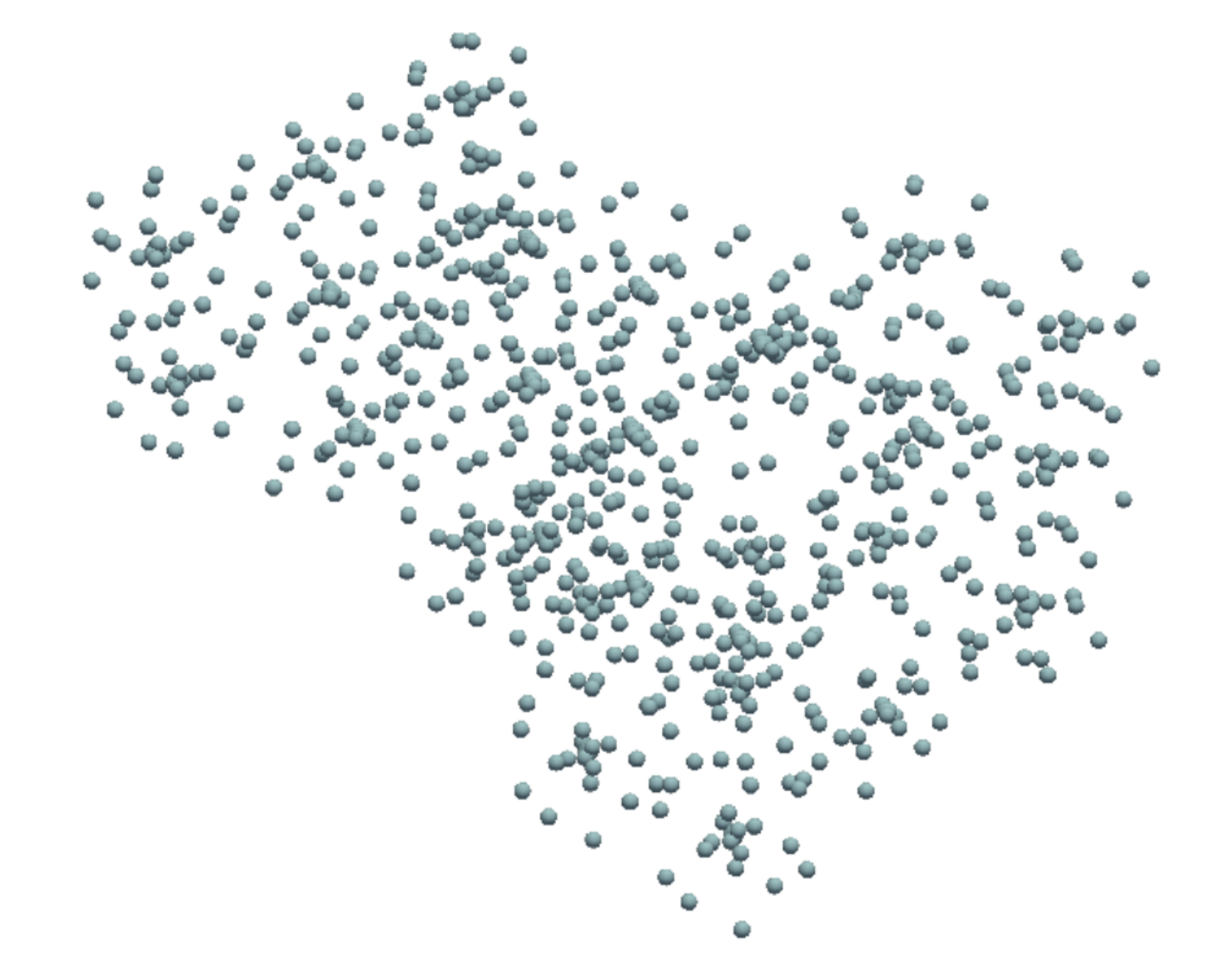} 
        \caption*{D-PCC (bpp: 0.281)}
    \end{subfigure}
    \hfill
    \begin{subfigure}{0.49\columnwidth}
        \centering
        \includegraphics[width=.8\linewidth]{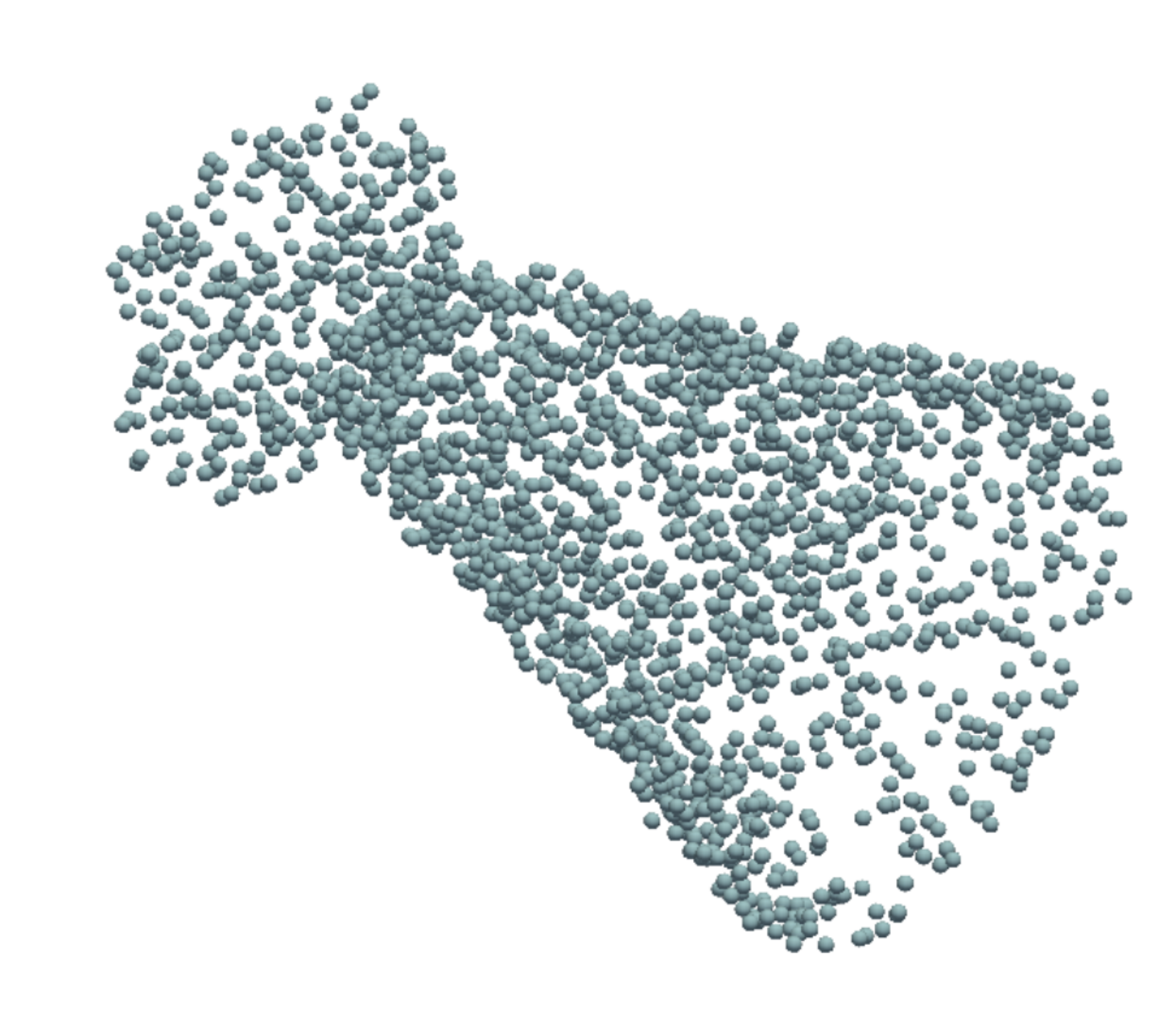} 
        \caption*{Ours (bpp: 0.125)}
    \end{subfigure}
    \caption{Low bit-rates PCC and compression artifacts.
    }
    \label{fig:teaser}
    \vspace{-1.5em}
    
\end{figure}

%% file: sec/2_related.tex
\section{Background on DDPM}
\label{sec:background}
Denoising Diffusion Probabilistic Models (DDPM) are a class of generative methods adopting a diffusion process to generate data. This process is defined as a Markov chain and allows the generation of high-fidelity samples of complex distributions. In a nutshell, given a sample from the target distribution $\mathbf{x}^{(0)}$, the \textit{diffusion} process gradually adds noise to $\mathbf{x}^{(0)}$. The \textit{reverse} process, indeed, involves a neural network to remove noise for generating a sample.
More formally, the diffusion process (or forward process) is defined as

\begin{equation}
\label{eq:forward}
    q(\mathbf{x}^{(1:T)} | \mathbf{x}^{(0)} ) = \prod_{t=1}^{T} q(\mathbf{x}^{(t)} | \mathbf{x}^{(t-1)}) ,
\end{equation}
where
\begin{equation}
\label{eq:for_gaussian}
    q(\mathbf{x}^{(t)} | \mathbf{x}^{(t-1)}) = \mathcal{N}\left(\mathbf{x}^{(t)}; \sqrt{1-\beta_{t}} \mathbf{x}^{(t-1)}, \beta_{t} \mathbf{I} \right)
\end{equation}
having $\mathbf{x}^{(t)}$ sampled from a Gaussian distribution with mean $\sqrt{1-\beta_{t}} \mathbf{x}^{(t-1)}$ and variance $\beta_{t} \mathbf{I}$, where $\beta_{t}$ is the variance scheduler and $\mathbf{I}$ is the identity matrix. For generating a latent sample $\mathbf{x}^{(t)}$, this process would require to perform $t-1$ steps in the Markov chain. To avoid this, following~\cite{ddpm}, we can express $\mathbf{x}^{(t)}$ directly in terms of $\mathbf{x}^{(0)}$:
\begin{equation}
\label{eq:for_gaussian_x0}
    q(\mathbf{x}^{(t)} | \mathbf{x}^{(0)}) = \mathcal{N}\left(\mathbf{x}^{(t)}; \sqrt{\bar{\alpha}_{t}} \mathbf{x}^{(0)}, (1-\bar{\alpha}_{t}) \mathbf{I} \right),
\end{equation}
where $\alpha_{t} = 1-\beta_{t}$ and $\bar{\alpha}_{t} = \prod_{s = 1}^{t} \alpha_s$.

For large $T$, the forward process converges to an isotropic Gaussian distribution centered in zero. The reverse process samples from this distribution and, adopting a neural network parameterized with $\theta$, gradually denoises the Gaussian noise into a real sample from the target distribution. Specifically, this process is defined as:
\begin{equation}
\label{eq:reverse}
    p_{\theta}(\mathbf{x}^{(0:T)} ) = p(\mathbf{x}^{(T)}) \prod_{t=1}^{T} p_{\theta}(\mathbf{x}^{(t-1)} | \mathbf{x}^{(t)}) ,
\end{equation}
where
\begin{equation}
    p_{\theta}(\mathbf{x}^{(t-1)} | \mathbf{x}^{(t)})  = \mathcal{N}(\mathbf{x}^{(t-1)}; \mathbf{\mu}_{\theta}(\mathbf{x}^{(t)}, t), \beta_{t} \mathbf{I} ),
\end{equation}
and $\mu_{\theta}$ is a network that predicts the mean of the Gaussian distribution for $\mathbf{x}^{(t-1)}$.

Ho~\etal~\cite{ddpm} proposed a tractable way for training the reverse process. In particular, they minimized the Mean Squared Error between the noise added in the forward process and the noise predicted by the model:

\begin{figure*}[t]
     \centering
    \includegraphics[width=.85\textwidth]{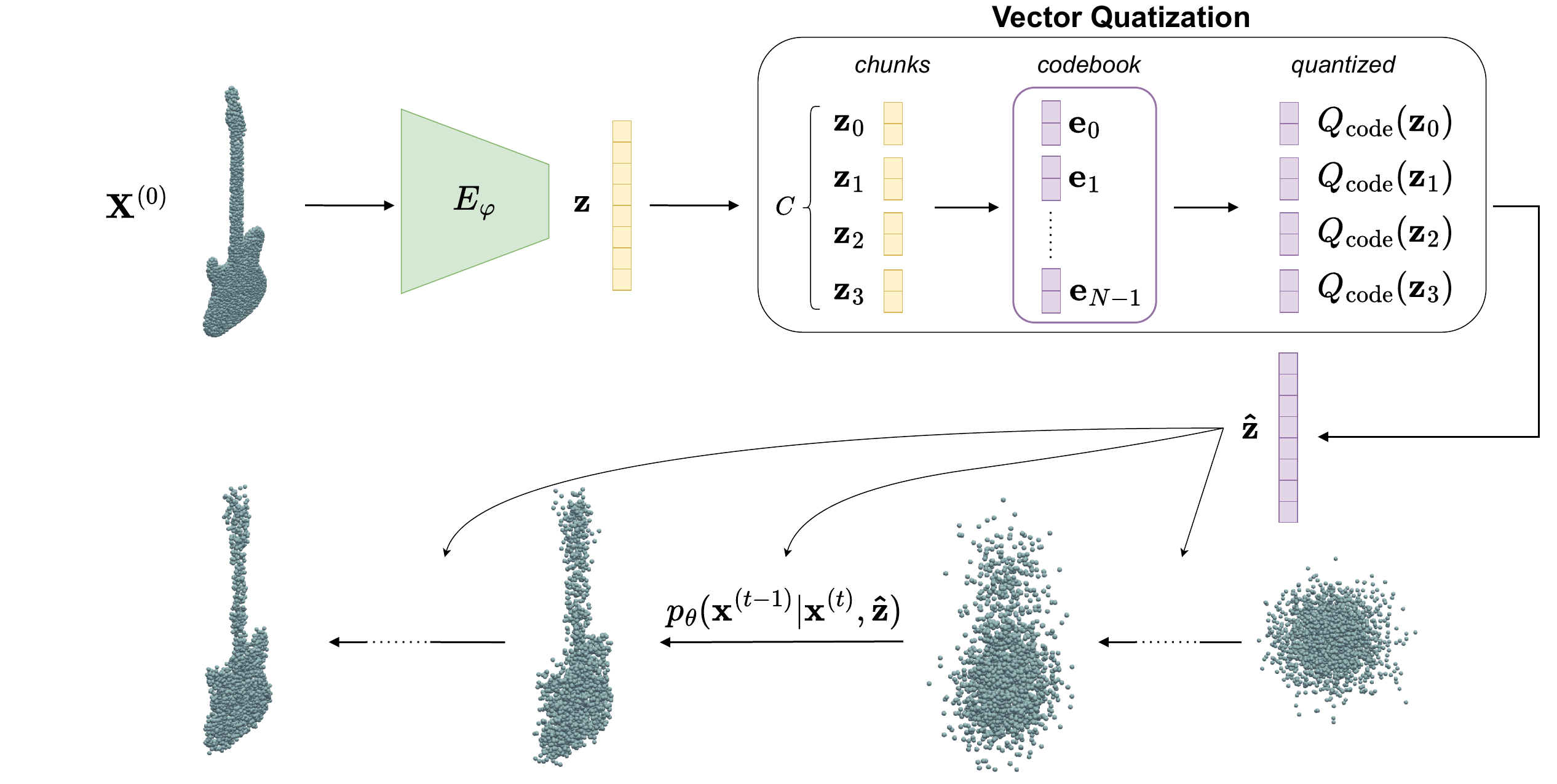}
    
    \caption{Overview of the proposed method. Here for simplicity, $\mathbf{x}^{(t-1)}$ represents a single point, while the diffusion process shows the final reconstructed point cloud (conditioned by the quantized latent representation $\mathbf{\hat{z}}$). Here $\mathbf{z}$ is split into $C = 4$ chunks, making $\mathbf{\hat{z}}$ representable with only $4$ codebook indices.} 
    \label{fig:overview}
    \vspace{-1em}
    
\end{figure*}

\begin{equation}
\label{eq:loss_diff}
\begin{split}
\mathcal{L}_{\text{Diff}} &= \mathbb{E}_{t,\mathbf{x}^{(0)},\mathbf{\epsilon}}  [ \Vert \mathbf{\epsilon} - \mathbf{\epsilon}_{\theta} (\sqrt{\bar{\alpha}_{t}} \mathbf{x}^{(0)} + \sqrt{1-\bar{\alpha}_{t}} \mathbf{\epsilon}, t) \Vert_2^{2} ]\\
&=  \mathbb{E}_{t,\mathbf{x}^{(0)},\mathbf{\epsilon}} [ \Vert \mathbf{\epsilon} - \mathbf{\epsilon}_{\theta} (\mathbf{x}^{(t)}, t) \Vert_2^{2} ],
\end{split}
\end{equation}
where $t \sim U(1, T)$, $\mathbf{\epsilon} \sim \mathcal{N}(0,\mathbf{I})$ is the noise and $\mathbf{\epsilon}_{\theta}(\cdot)$ is the trainable network that predicts the added noise. 
\\[.5em]
\noindent
\textbf{Point Cloud Generation. } Lou~\etal~\cite{wei} proposed a DDPM-based model for 3D Point Cloud Generation. More specifically, they considered a point cloud $\mathbf{X}^{(0)} = \{ \mathbf{x}_{i}^{(0)}\}_{i=1}^{P}$ as a set of $P$ points in an evolving system. Each of these points $\mathbf{x}_{i}$ can then be treated as independently sampled from a point distribution. For this reason, the forward diffusion process and the reverse diffusion one (detailed in \eqref{eq:forward} and \eqref{eq:reverse} respectively) are directly applied independently to these points.

In our work, we show how such modeling can be leveraged towards the distinct problem of compressing point clouds at very low bit rates.

%% file: sec/3_method.tex
\section{Proposed Method}
\label{sec:method}
In the previous section, we provided some background knowledge regarding Denoising Diffusion Probabilistic Models (DDPM). Here, we explain how this class of generative models can be employed as an autoencoder trained for a compression task. Next, we discuss the compression strategy employed to quantize the latent space and finally, we present the overall proposed framework.

\subsection{Diffusion-based Point Cloud Compression Model}
\label{subsec:comp_model}
To perform compression, our generative diffusion model needs to be conditioned during the reverse process to generate a specific sample. Moreover, each sample from the data distribution is represented as point cloud $\mathbf{X}^{(0)} \in \mathbb{R}^{P \times 3}$. 

Leveraging the point cloud auto-encoder proposed by~\cite{wei} we generate the original point cloud from a random Gaussian distribution in a point-to-point way. Specifically, to obtain certain sample $\mathbf{X}^{(0)}$, the generation is guided with a latent representation of the sample $\mathbf{z} = E_{\varphi}(\mathbf{X}^{(0)})$, where in our case $E_{\varphi}(\cdot)$ is a PointNet-based~\cite{pointnet} encoder having parameters $\varphi$. Next, we adopt the reverse diffusion process as a decoder, conditioned on the latent representation $\mathbf{z}$. To do this, we can define $ p_{\theta}(\mathbf{x}^{(t-1)} | \mathbf{x}^{(t)}, \mathbf{z})$ in~\eqref{eq:reverse} and rewrite the Loss function~\eqref{eq:loss_diff} training the auto-encoder to minimize: 

\begin{equation}
\mathcal{L}_{\text{Diff}} =  \mathbb{E}_{t,\mathbf{X}^{(0)},\mathbf{\epsilon}} [ \Vert \mathbf{\epsilon} - \mathbf{\epsilon}_{\theta} (\mathbf{x}^{(t)}, t, E_{\varphi}(\mathbf{X}^{(0)})) \Vert_2^{2} ].
\end{equation}

At this point, by appropriately quantizing and then compressing the latent representation $\mathbf{z}$, we obtain a model able to compress and then generate the input point cloud. In the next section, we describe how we compress this latent code.

\subsection{Encoding the Latent Representation}
\label{subsec:econde_latent}
The input point cloud $\mathbf{X}^{(0)}$ is encoded into the latent representation vector $\mathbf{z} \in \mathbb{R}^{d}$, where $d$ is the hidden dimension. Inspired by VQ-VAE~\cite{vq}, we adopt a vector quantization strategy based on a learnable codebook to efficiently compress our latent code. Specifically, to control the rate, we divide $\mathbf{z}$ into $C$ chucks: $\mathbf{z} = \{\mathbf{z}_{0}, \dots , \mathbf{z}_{C-1} \}$, having the same dimension $\mathbf{z}_{i} \in \mathbb{R}^{d / C}$. We choose $C$ as an integer multiple of $d$. Then, the $i$-th chunk $\mathbf{z}_{i}$ is mapped to a vector in the learned codebook minimizing the vector quantization loss:

\begin{equation}
\mathcal{L}_{\text{VQ}} =  \mathbb{E}_{\mathbf{z}_{i}} [ \Vert \sg(\mathbf{z}_{i}) - \mathbf{e}_{q} \Vert_2^{2} + \Vert \sg(\mathbf{e}_{q}) - \mathbf{z}_{i} \Vert_2^{2} ] ,
\end{equation}
where $\sg(\cdot)$ stands for the stopgradient operator that is defined having zero partial derivatives, while $\mathbf{e}_{q}$ is the closest codebook entry to $\mathbf{z}_{i}$, defined as $\mathbf{e}_{q} = Q_{\text{code}}(\mathbf{z}_{i})$. This loss ensures that the codebook entries match the encoder's output and, at the same time, forces the encoder to produce latent vectors close to the codebook entries. 

The quantized version of the latent code $\mathbf{\hat{z}}$ is given by mapping each of its chunks with the closest entries in the codebook $\mathbf{\hat{z}} = \{Q_{\text{code}}(\mathbf{z}_{0}), \dots, Q_{\text{code}}(\mathbf{z}_{C-1})  \}$. 

At this point, the rate of $\mathbf{\hat{z}}$ can be easily defined in terms of $C$ and the codebook dimension. In particular, $\mathbf{\hat{z}}$ can be represented by specifying the indexes of $C$ entries in the codebook. The number of bits per index depends on the size of the codebook. For instance, if the codebook has a size of $N = 256$, each index requires 8 bits for representation. Therefore, the final rate can be computed as $C \times \log_{2} N$.

\subsection{Overall Framework}
\label{subsec:overall}
In our framework, the conditional diffusion model and the learnable vector quantizer are jointly trained minimizing: 
\begin{equation}
    \mathcal{L} = \mathcal{L}_{\text{diff}} + 
    \mathcal{L}_{\text{VQ}}.
\end{equation}
Notably, the $\mathcal{L}_{\text{diff}}$ adopted here differs from \eqref{eq:loss_diff} since $E_{\varphi}(\mathbf{X}^{(0)}) = \mathbf{z}$ is replaced with $\mathbf{\hat{z}}$.
This joint optimization enables better integration of the quantization process with the generative model.

Fig.~\ref{fig:overview} shows an illustrative overview of our compression pipeline once the model is trained. Here, the point cloud to be compressed is first passed through the encoder $E_{\varphi}(\cdot)$, producing a latent representation $\mathbf{z}$. This vector is then divided into $C$ chunks, and each chunk is quantized using the learned codebook to produce $\mathbf{\hat{z}}$. The latter is used to condition the reverse denoising process of the diffusion model to reconstruct the input point cloud, which thus functions as a true decoder.

As discussed in Sec.~\ref{subsec:econde_latent}, the final rate depends on the number of chunks $C$ and the codebook size $N$. For simplicity, we set $N = 128$ constant. Then, to change the coding rate, we can use different $C$ values from $1$ up to $d$ (where $\mathbf{z} \in \mathbb{R}^{d}$). Intuitively, with $C = d$, each value in $\mathbf{z}$ is independently quantized, leading smaller quantization error. On the contrary, with $C = 1$, the entire vector is quantized with a single index, resulting in only $N$ possible reconstructions. Is important to notice that each codebook has to be constructed based on the chunk dimension: $d/C$. For this reason, given $C$ and thus the target rate, the corresponding codebook is trained with the entire model and then used for coding at that rate.

%% file: sec/4_experiments.tex
\input{plots/rd}

\section{Experiments}
\label{sec:experiments}
In this section, we detail the experimental setup and the results obtained with our compression method. Lastly, we present two ablation
studies on key design choices: the compression method and the encoder architecture. 
\subsection{Experimental Setup}
\noindent
\textbf{Datasets. } We conduct our experiments on ShapeNet~\cite{shapenet} and ModelNet40~\cite{modelnet}. ShapeNet~\cite{shapenet} contains a total of $51,127$ shapes, divided in $55$ categories. This dataset was randomly split to obtain $43,433$ samples for training and $7,694$ samples for testing. We sample $2,048$
points from each shape to acquire the point clouds
and we normalize each of them to zero mean and unit variance.
ModelNet40~\cite{modelnet} is a dataset containing $12,308$ mashes CAD model from $40$ categories. From this dataset, we adopted $9,843$ samples for training and the remaining $2,468$ for testing. Also in such case, we sample $2,048$ points from each mesh and we normalized the coordinates in the same manner.
\\[.5em]
\noindent
\textbf{Training Details. } Following the same training procedure adopted in~\cite{wei}, we trained our compression model for around $1,000,000$ steps, using a batch of $128$ samples for each step. We used Adam as optimizer~\cite{adam} with an initial learning rate of $1 \times 10^{-4} $ and a final learning rate of $1 \times 10^{-5}$, decreased linearly from step $200k$ to $400k$.
\input{plots/qualitative}
Regarding the model, we choose a dimension $d = 256$ for the encoder's output.
As implemented in~\cite{wei}, we used $T = 200$ denoising steps and a linear variance schedule from $\beta_{1} = 0.0001$ to $\beta_{T} = 0.05$.

As detailed in Sec.~\ref{subsec:econde_latent}, we obtained a pool of models to compress the input point cloud at different bit-rates by using chunks numbers $C$ belonging to $\{4,8,16,32,64,128,256\}$ (lowering the $C$ lowers the rate).  For each of these $C$ values, a specific codebook is jointly trained with the autoencoder to quantize chucks of dimension $d/C$. For simplicity, we fix the codebook size $N = 128$.
\\
\noindent
\textbf{Evaluation. }We compare our method with both non-learning-based techniques: G-PCC (release-v23.0)~\cite{gpcc}, Google Draco~\cite{draco}; and learning-based models: D-PCC~\cite{dpcc}, COT-PCC~\cite{cot_pcc}. For these latter models, we re-trained them on the same datasets as our method, using $\lambda$ values in the Rate-Distortion Loss function to achieve a low bit-rate compression.

Following~\cite{dpcc, cot_pcc}, we evaluate the accuracy in geometry reconstruction in terms of symmetric point-to-point Chamfer Distance (CD) and point-to-plane PSNR. We also compare the methods by leveraging the Earth Mover’s distance (EMD), while the compression rate is measured in Bits per Point (bpp).

\begin{figure}[t]
    \centering
    \includegraphics[width=.75\linewidth]{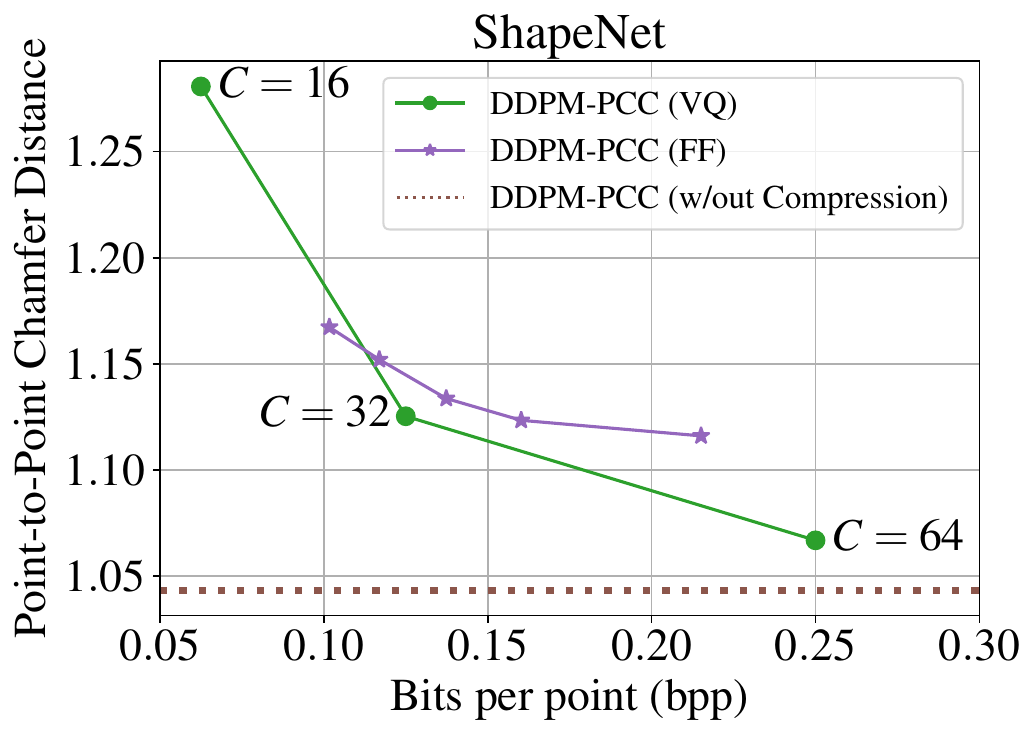}
    \caption{Comparison between a learnable vector quantize approach (VQ) and a fully factorize-based strategy (FF) to encode the latent representation. The dashed line represents the lower bound in Chamfer Distance: the generative model. \\Different values of $C$ allow for different RD trade-offs}
    \label{fig:ablation_compression}
    \vspace{-.5em}
    
\end{figure}

\subsection{Rate-Distortion Comparison}
\label{subsec:rd}
\noindent
\textbf{Quantitative Results. } We first provide quantitative results by comparing our method with previous state-of-the-art approaches. In Fig.~\ref{fig:rd} we show the Chamfer Distance (CD), the PSNR, and Earth Mover’s distance (EMD) of all methods against Bits per point (bpp). To highlight differences among the methods, the y-axes are plotted using a logarithmic scale. 

The obtained results show that our method achieves superior performance compared to the other analyzed approaches on both datasets and according to all quality metrics. In particular, the advantages of our method are particularly pronounced at low bit rates, where it provides significantly better reconstruction quality.
Draco~\cite{draco}, instead, requires a large number of bits to achieve acceptable reconstruction. 
Although G-PCC~\cite{gpcc} shows good capabilities in preserving the overall geometry, it suffers from a loss of finer details due to the voxelization process. Finally, the learnable methods~\cite{dpcc, cot_pcc} do not converge to satisfactory results for the compression levels analyzed, as they were designed to operate at higher bit-rates (typically in the range of $1$ to $6$ bpp).

Despite these advantages, our method is limited by the performance of the generative model. In fact, even by increasing the coding rate, the results obtained cannot outperform the quality of reconstruction achieved by the generative model.
\\[.5em]
\noindent
\textbf{Qualitative Results. } To clearly show the reconstruction ability of our compression model, even at a very low bit-rate (around $0.065$ bpp), in Fig.~\ref{fig:qualitative} we report some qualitative results at various bit-rates. From these reconstructions, we can notice the limitations encountered by previous techniques in such a context. Here, G-PCC~\cite{gpcc} shows the effect of the voxelization and quantization needed to compress at these rates. This leads to the loss of many local details during reconstruction. D-PCC~\cite{dpcc}, indeed, consistently requires more bits for low-quality reconstructions. On the other hand, our model can achieve satisfactory PSNR results even for very low bpp in all reconstructions analyzed.

\subsection{Ablation Studies}
\noindent
\textbf{Compression Strategy. } 
In Fig.~\ref{fig:ablation_compression} we compare our chosen quantization technique, which exploits learnable vector quantizer (VQ), with a more traditional end-to-end compression approach~\cite{balle17}, where the latent space is modeled with a fully factorized (FF) distribution estimated by an auxiliary neural network.
In the latter, the compression is achieved by balancing (with a $\lambda$ parameter) the reconstruction and the rate of the latent representations in the well-known rate-distortion (RD) loss.
 From ths figure we notice that our method achieves better compression trade-offs, obtaining closer results to the model that does not implement any type of compression. Moreover, our approach does not involve the use of a $\lambda$ parameter to adjust the importance of reconstruction versus rate. 
 Indeed, as reported in Fig.~\ref{fig:ablation_compression}, by changing the value of $C$ we can easily tailor the model to a specific coding rate. 
\\[.5em]
\noindent
\textbf{Encoder Architecture. }  
Finally, we replace PointNet with PointNet++~\cite{pointnet++} concerning the encoder architecture for the generative model (\ie without  quantization). 
In Tab~\ref{tab:ablation_enc} we notice that PointNet++~\cite{pointnet++}, since it leverages the geometry of the input point cloud, can result in a richer latent representation. This increases the performance of the model in terms of CD, but at the cost of increasing coding time by $\times 100$. PointNet~\cite{pointnet}, indeed, considers each point independently. This results in a less informative representation but obtained in a much faster time.

\begin{table}[t]
\centering
\begin{tabular}{@{}lccc@{}}
\toprule
Encoder & Chamfer-Distance ($\downarrow$) & Enc. (s) & Dec. (s) \\
\midrule
PointNet~\cite{pointnet} & $1.043$ & $\mathbf{0.001_{\pm 0.001}}$ & $0.254_{\pm 0.0102}$ \\ 
PointNet++~\cite{pointnet++} & $\mathbf{0.953}$ & $0.134_{\pm 0.004}$ & $0.259_{\pm 0.012}$ \\ 
\bottomrule
\end{tabular}
\caption{
Ablation study concerning the encoder architecture. Encodings and Decondings times are in seconds and evaluated on NVIDIA GeForce RTX 3090 GPU.
}
\vspace{-2em}
\label{tab:ablation_enc}
\end{table}

%% file: plots/rd.tex
\begin{figure*}[t]
    \centering
    \begin{subfigure}{0.32\textwidth}
        \centering
        \includegraphics[width=\linewidth]{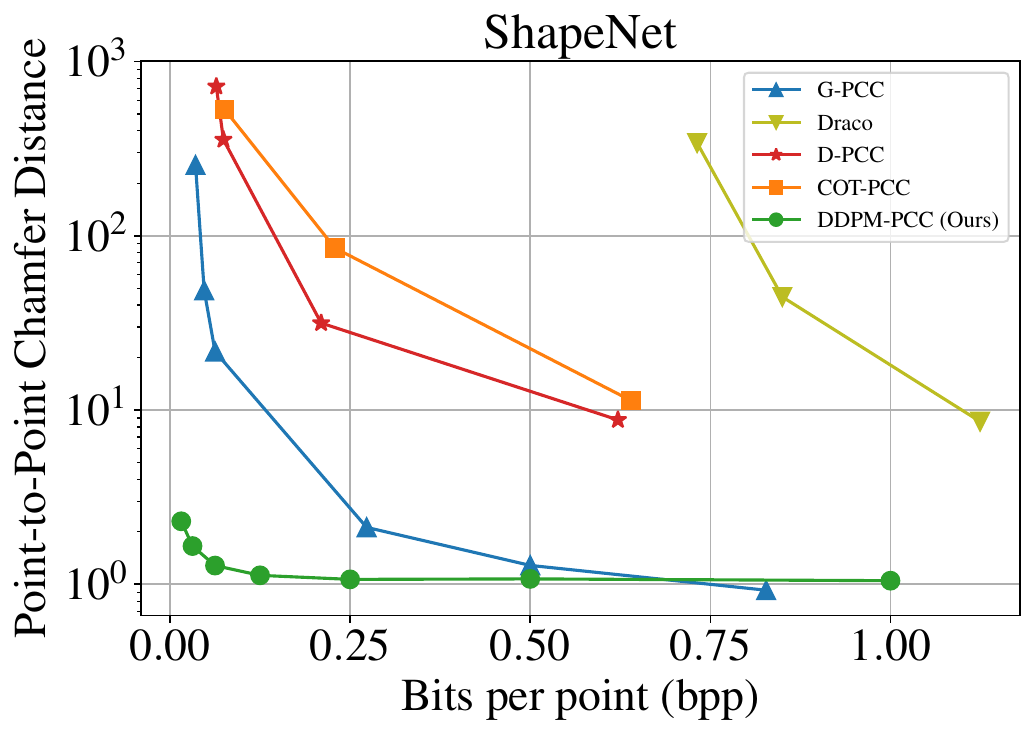} 
    \end{subfigure}
    \hfill
    \begin{subfigure}{0.32\textwidth}
        \centering
        \includegraphics[width=\linewidth]{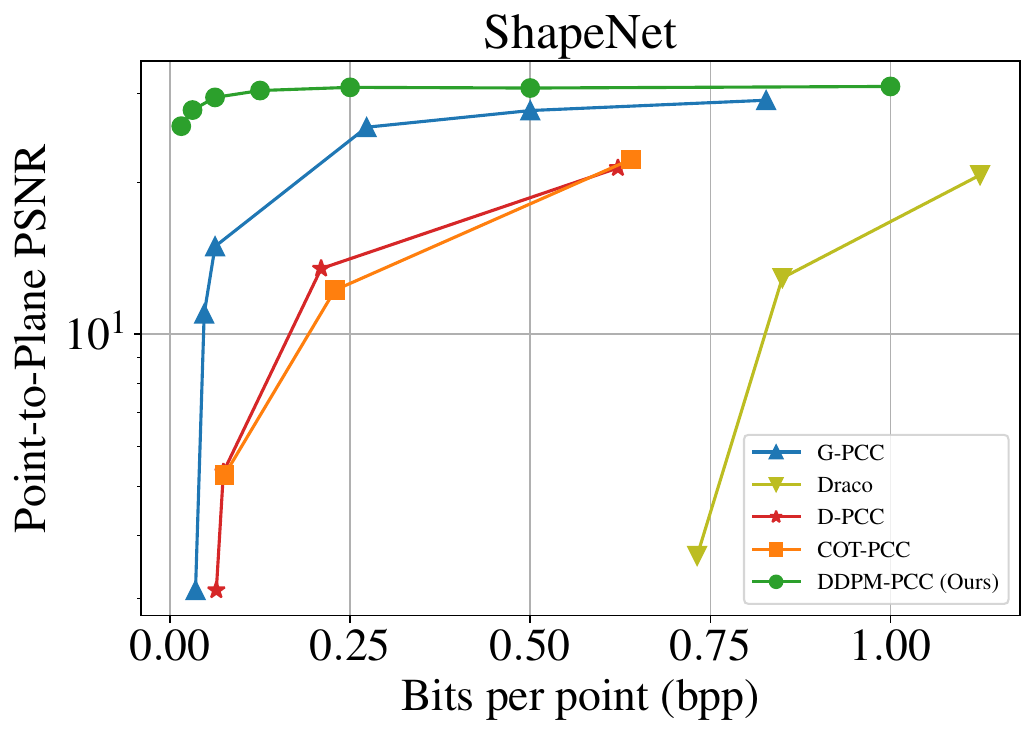} 
    \end{subfigure}
    \hfill
    \begin{subfigure}{0.32\textwidth}
        \centering
        \includegraphics[width=\linewidth]{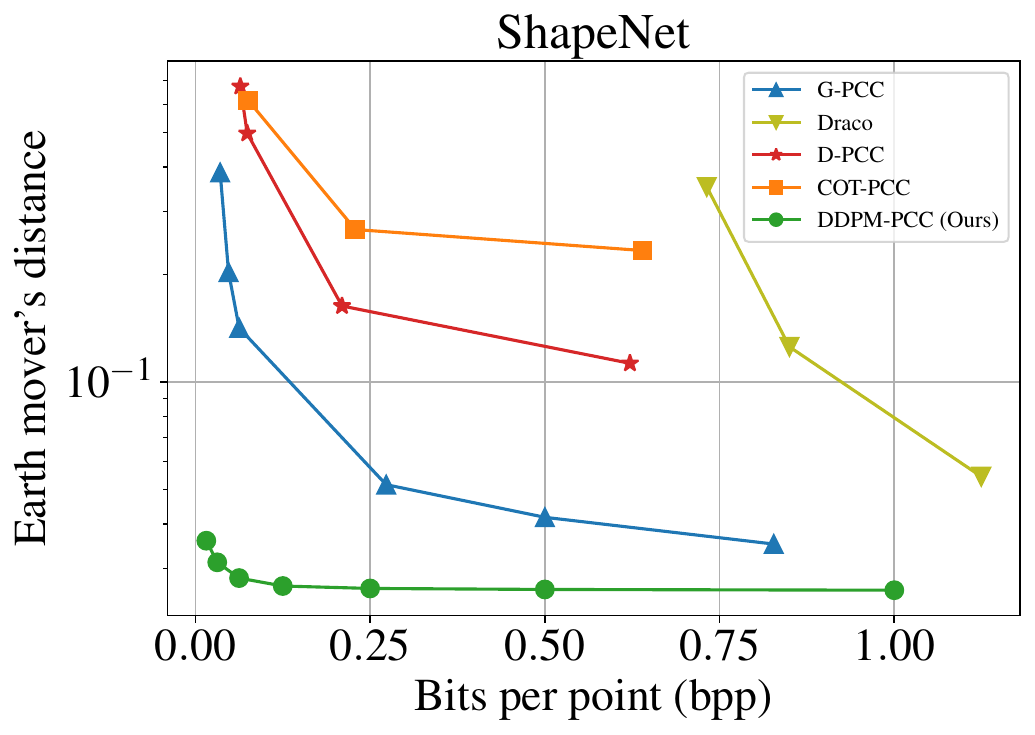} 
    \end{subfigure}
    
    \vspace{0.5cm}
    
    \begin{subfigure}{0.32\textwidth}
        \centering
        \includegraphics[width=\linewidth]{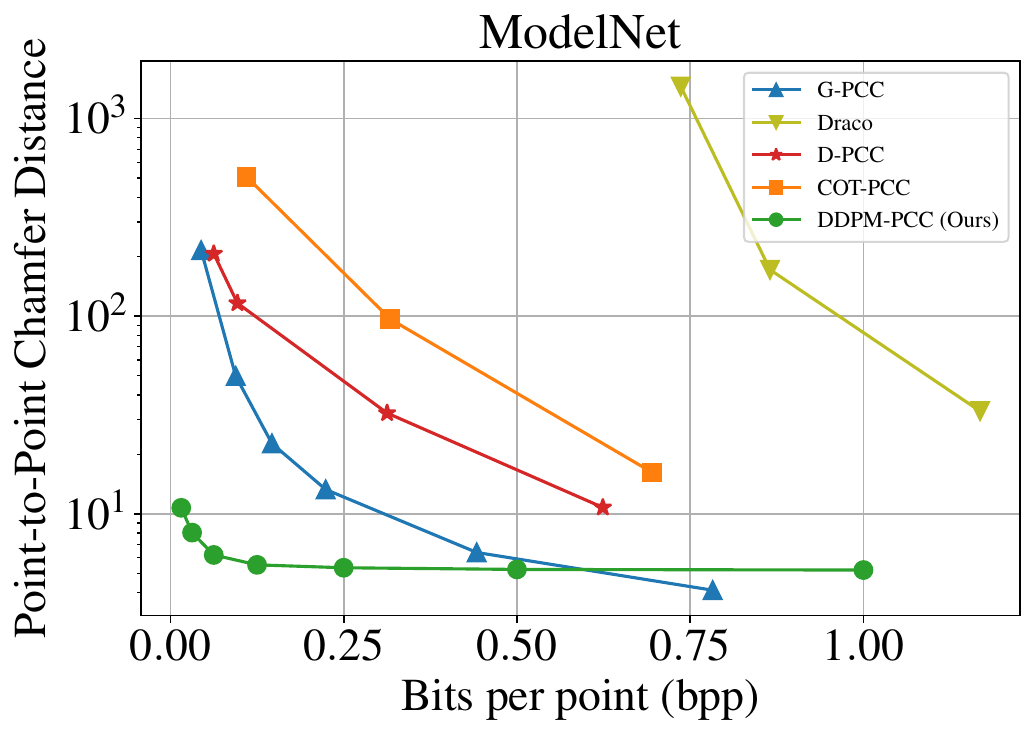} 
    \end{subfigure}
    \hfill
    \begin{subfigure}{0.32\textwidth}
        \centering
        \includegraphics[width=\linewidth]{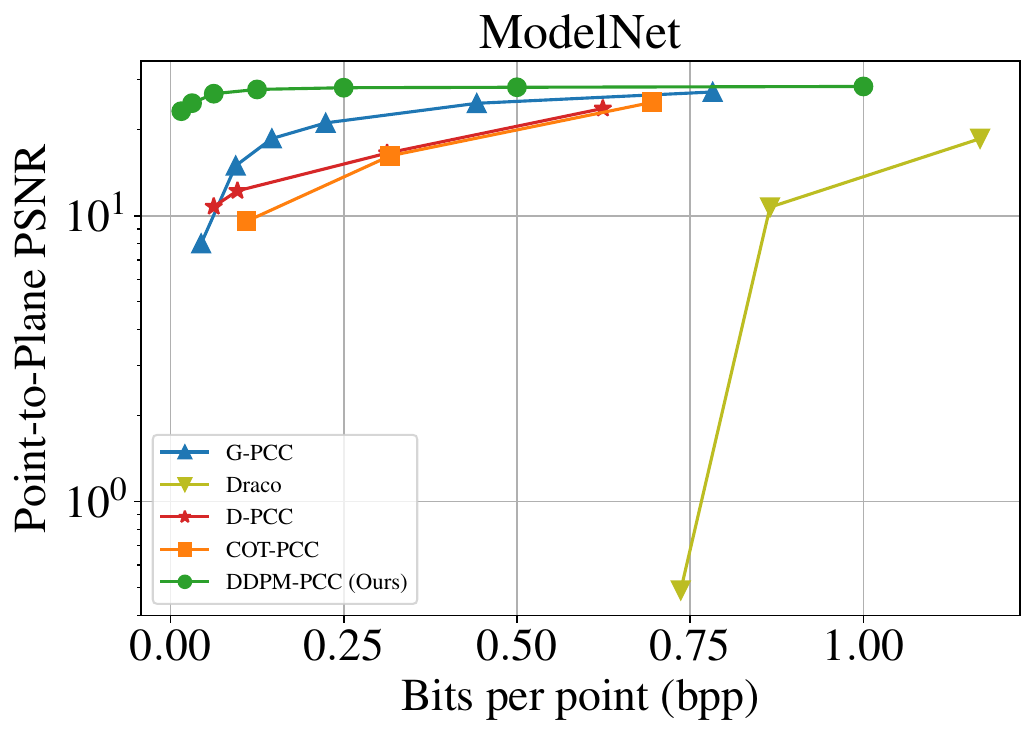} 
    \end{subfigure}
    \hfill
    \begin{subfigure}{0.32\textwidth}
        \centering
        \includegraphics[width=\linewidth]{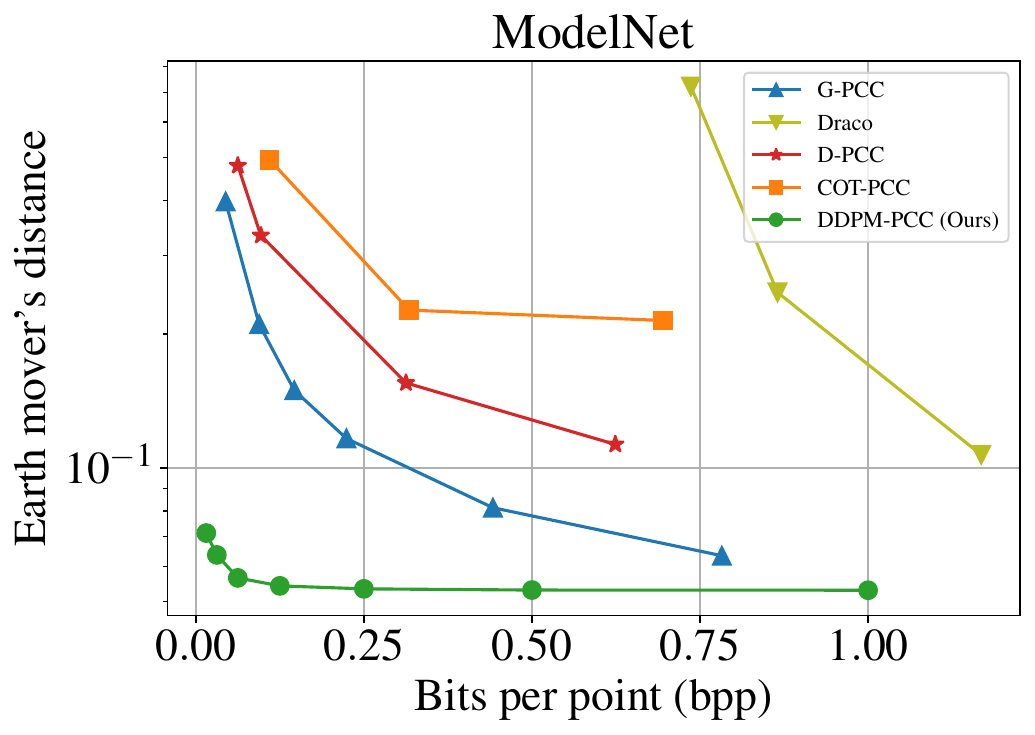} 
    \end{subfigure}
    
    \caption{Quantitative results on ShapeNet (first row) and ModelNet40 (second row). The y-axes are plotted in logarithmic scale. Our method achieves the best geometry reconstruction at a very low bit rate according to all quality metrics.}
    \label{fig:rd}
    \vspace{-1em}
    
\end{figure*}

%% file: plots/qualitative.tex
\begin{figure*}[t]
    \centering
    \makebox[0.22\textwidth]{\textbf{Ground Truth }}%
    \hfill
    \makebox[0.22\textwidth]{\textbf{G-PCC~\cite{gpcc}}}%
    \hfill
    \makebox[0.22\textwidth]{\textbf{D-PCC~\cite{dpcc}}}%
    \hfill
    \makebox[0.22\textwidth]{\textbf{DDPM-PCC (Ours)}}\\[1em]
    
    \begin{subfigure}{0.22\textwidth}
        \centering
        \includegraphics[width=.82\linewidth]{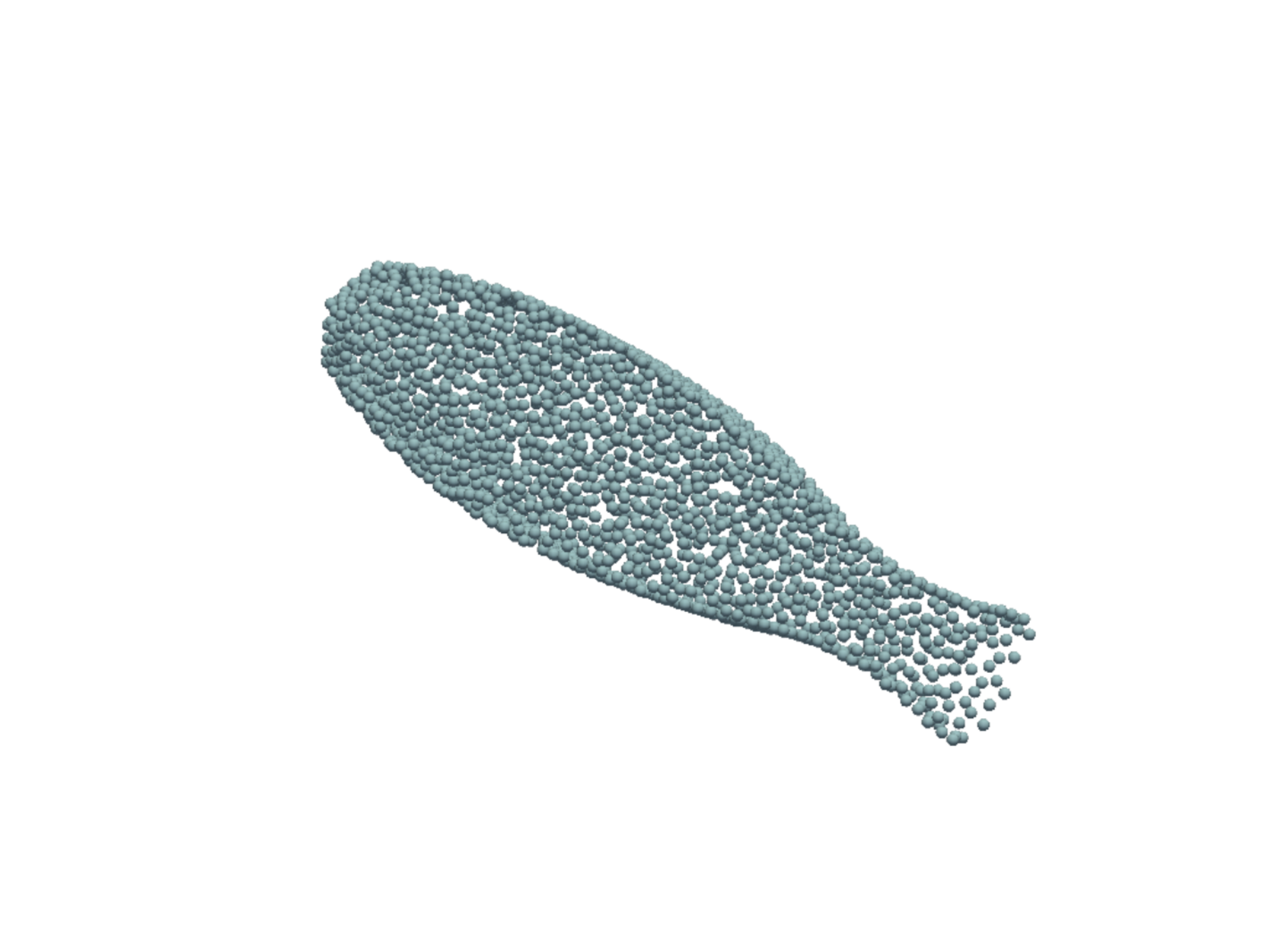} 
        \caption*{--}
    \end{subfigure}
    \hfill
    \begin{subfigure}{0.22\textwidth}
        \centering
        \includegraphics[width=.82\linewidth]{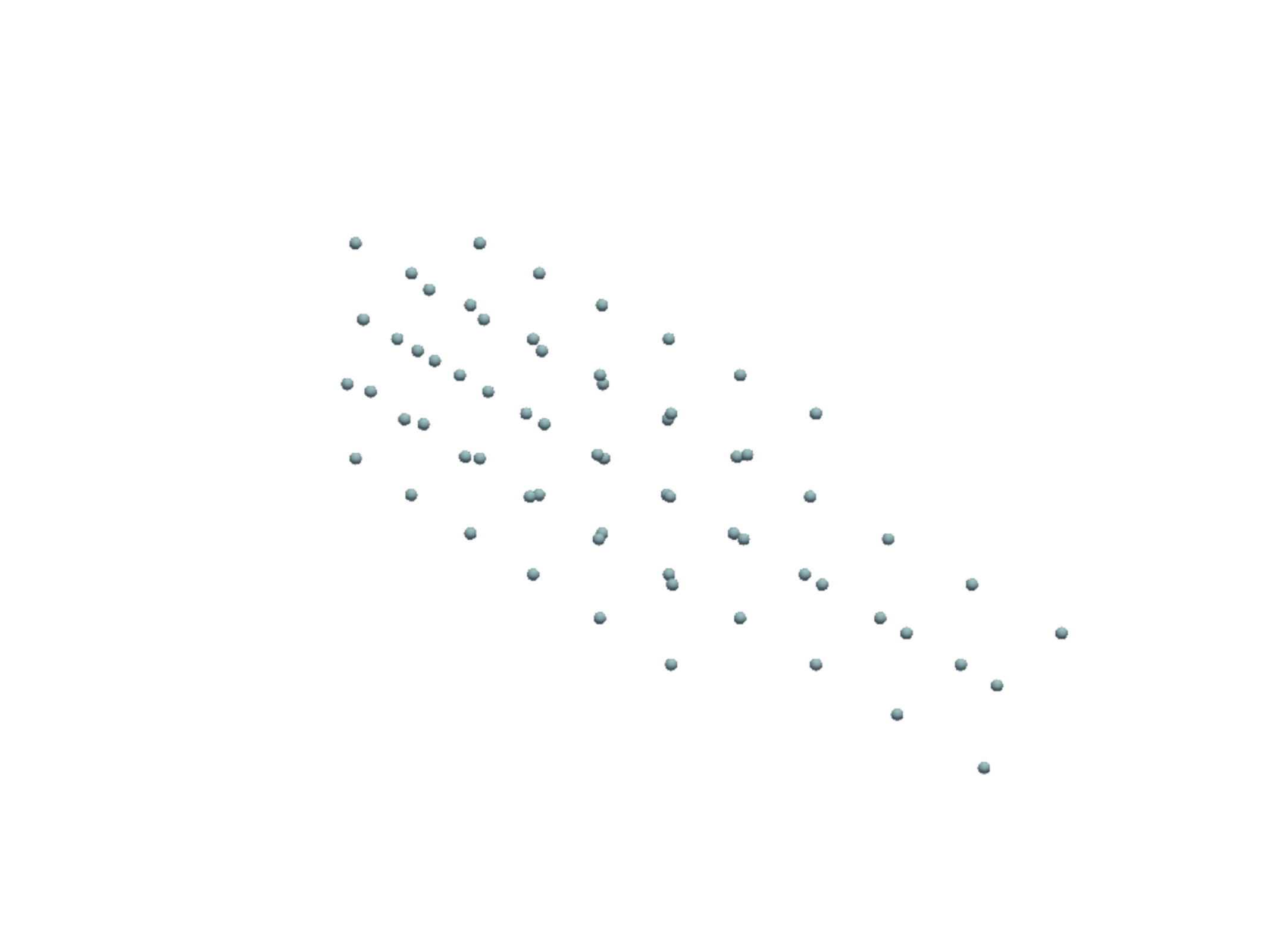} 
        \caption*{bpp: 0.10 - PSNR: 17.25}
    \end{subfigure}
    \hfill
    \begin{subfigure}{0.22\textwidth}
        \centering
        \includegraphics[width=.82\linewidth]{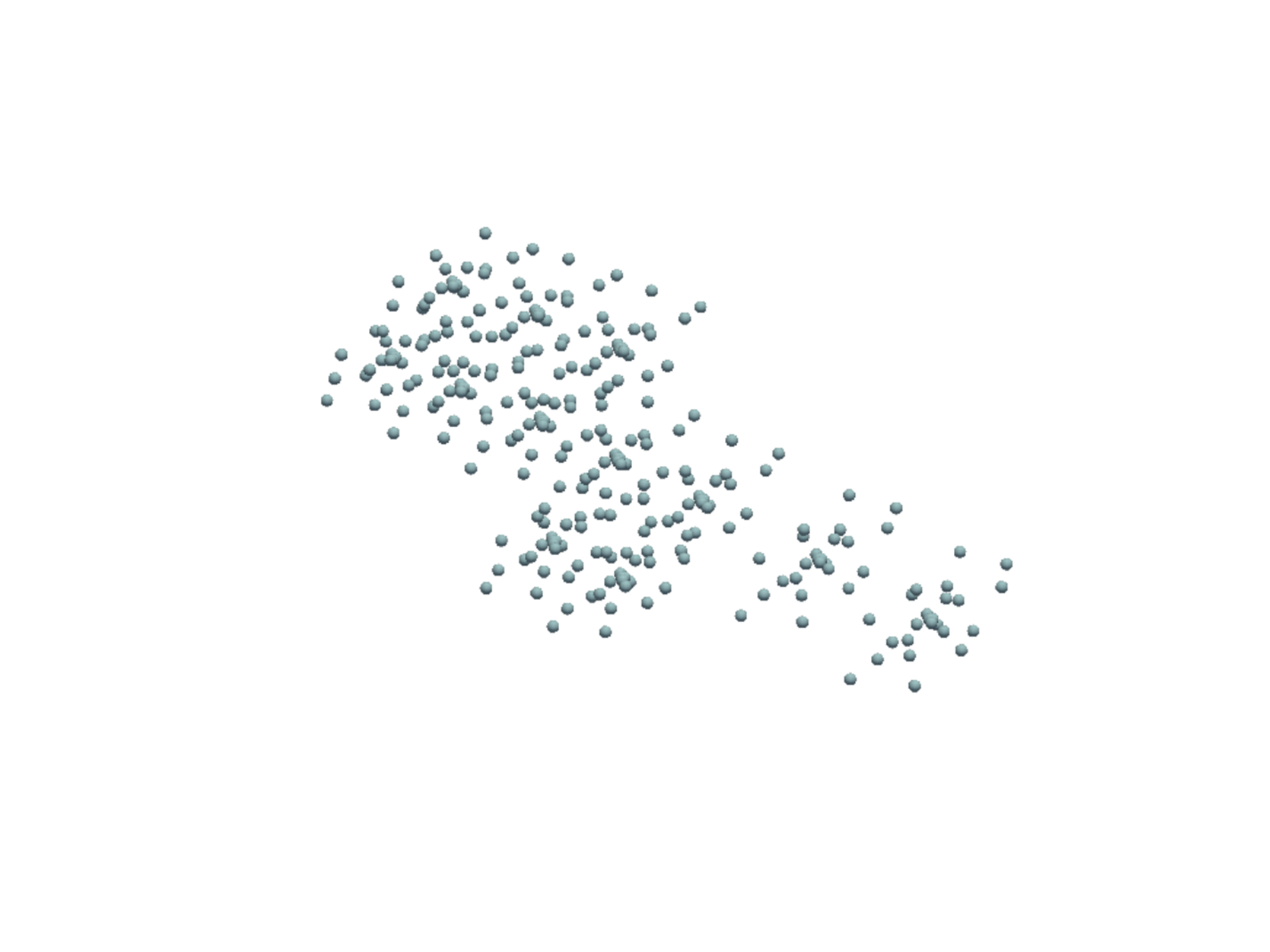} 
        \caption*{bpp: 0.25 - PSNR: 13.06}
    \end{subfigure}
    \hfill
    \begin{subfigure}{0.22\textwidth}
        \centering
        \includegraphics[width=.82\linewidth]{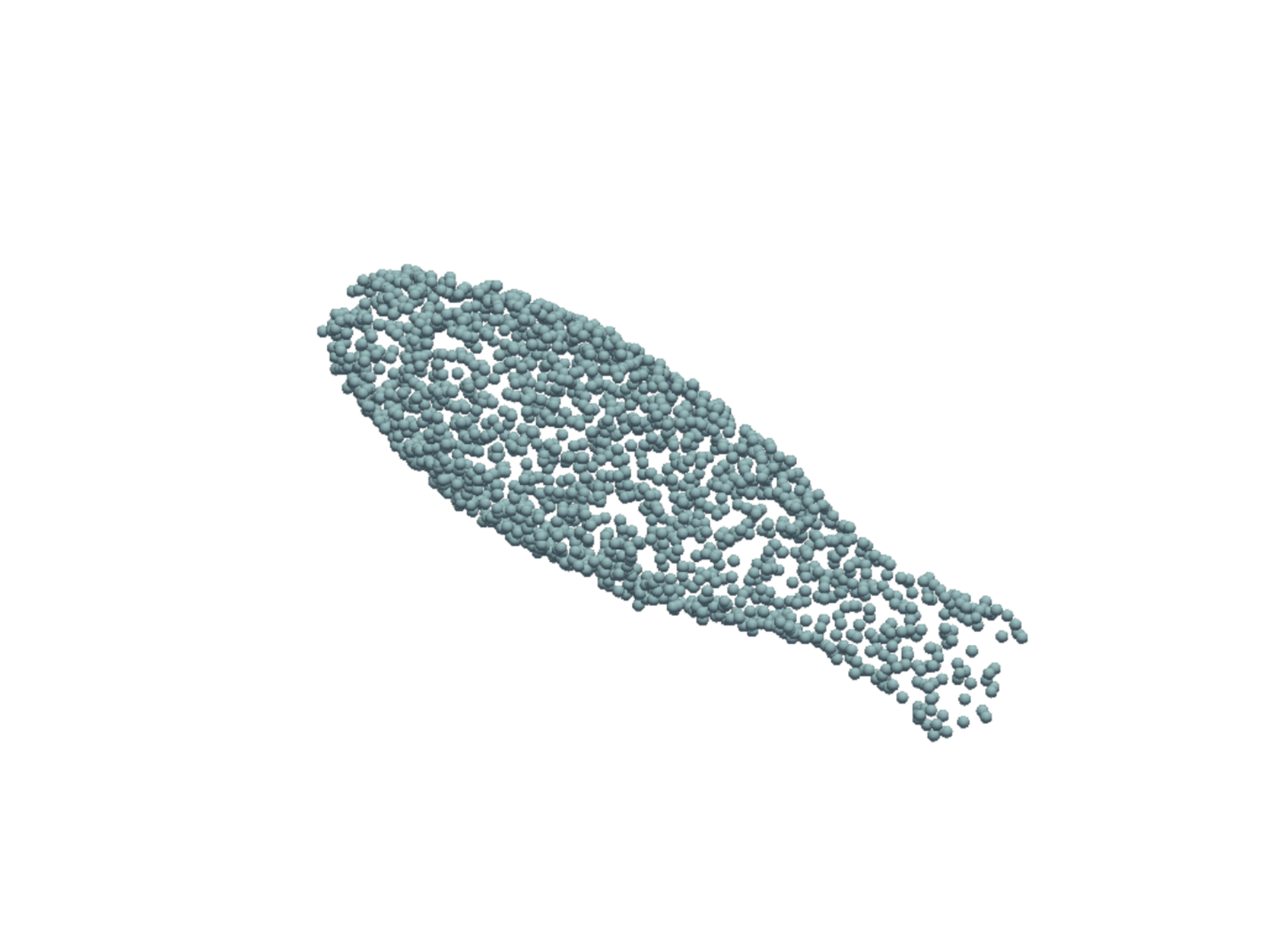} 
        \caption*{bpp: 0.065 - PSNR: 32.97}
    \end{subfigure}
    
    \vspace{0.5cm}
    
    \begin{subfigure}{0.22\textwidth}
        \centering
        \includegraphics[width=.82\linewidth]{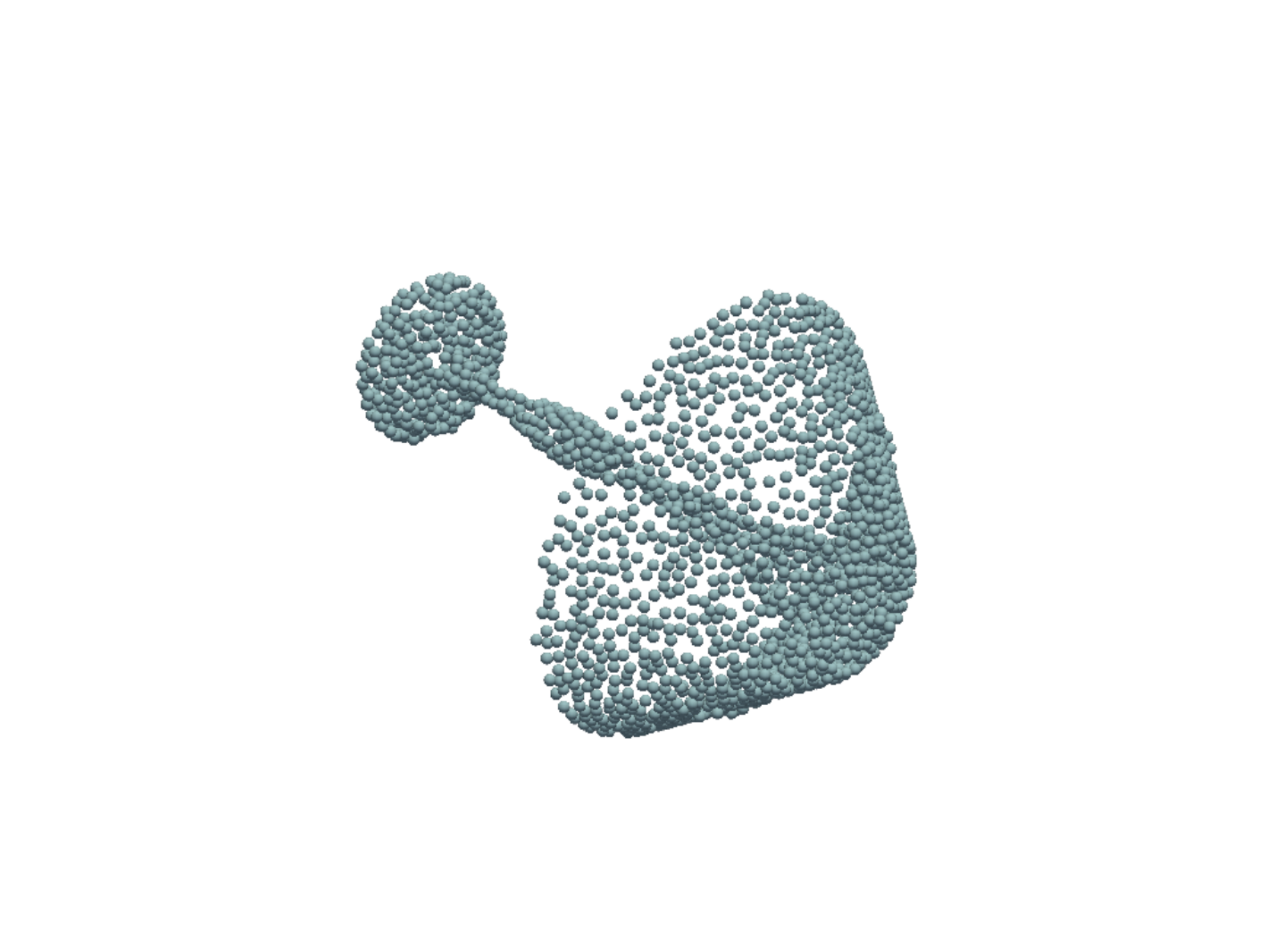} 
        \caption*{--}
    \end{subfigure}
    \hfill
    \begin{subfigure}{0.22\textwidth}
        \centering
        \includegraphics[width=.82\linewidth]{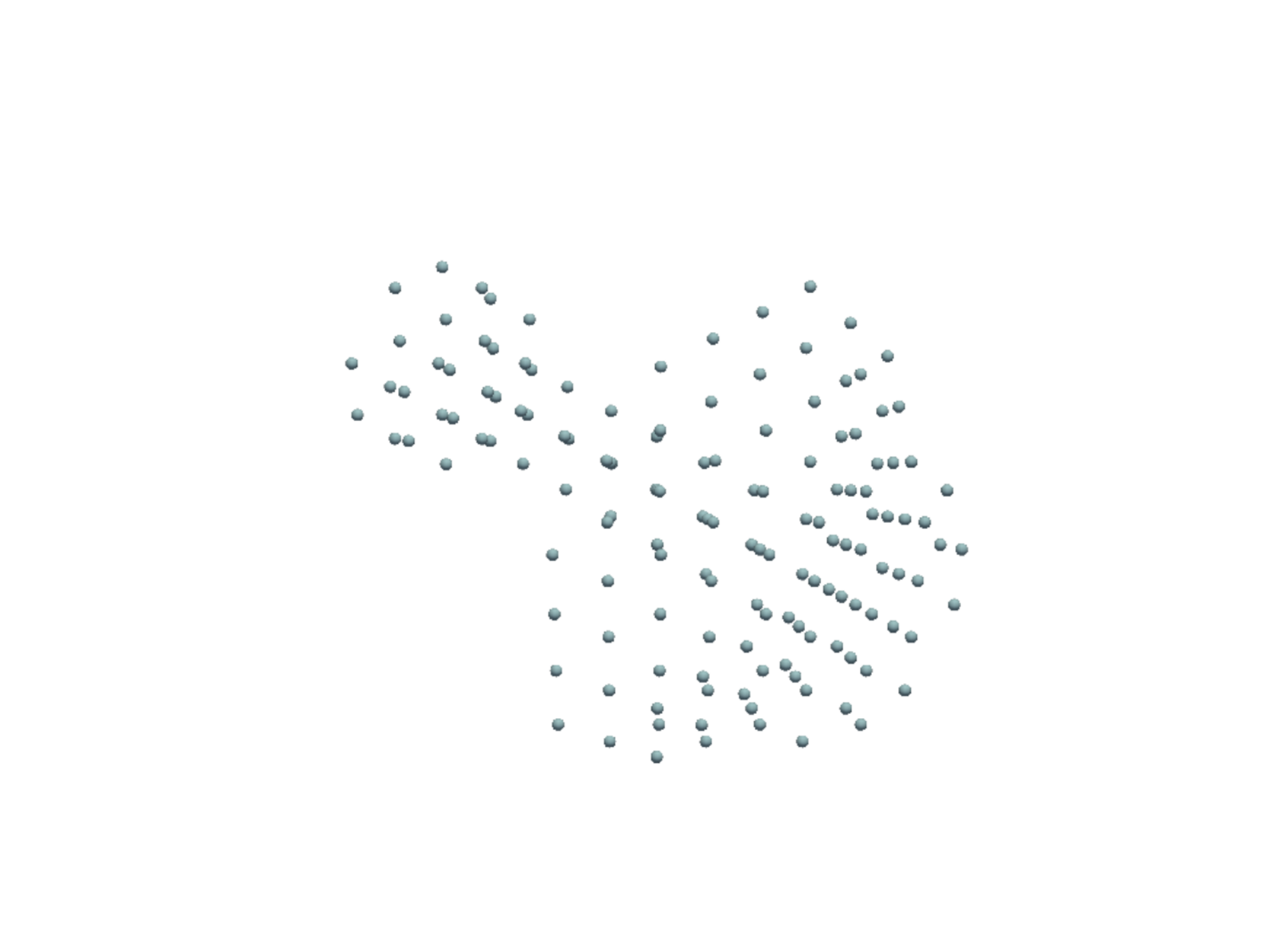} 
        \caption*{bpp: 0.25 - PSNR: 18.64}
    \end{subfigure}
    \hfill
    \begin{subfigure}{0.22\textwidth}
        \centering
        \includegraphics[width=.82\linewidth]{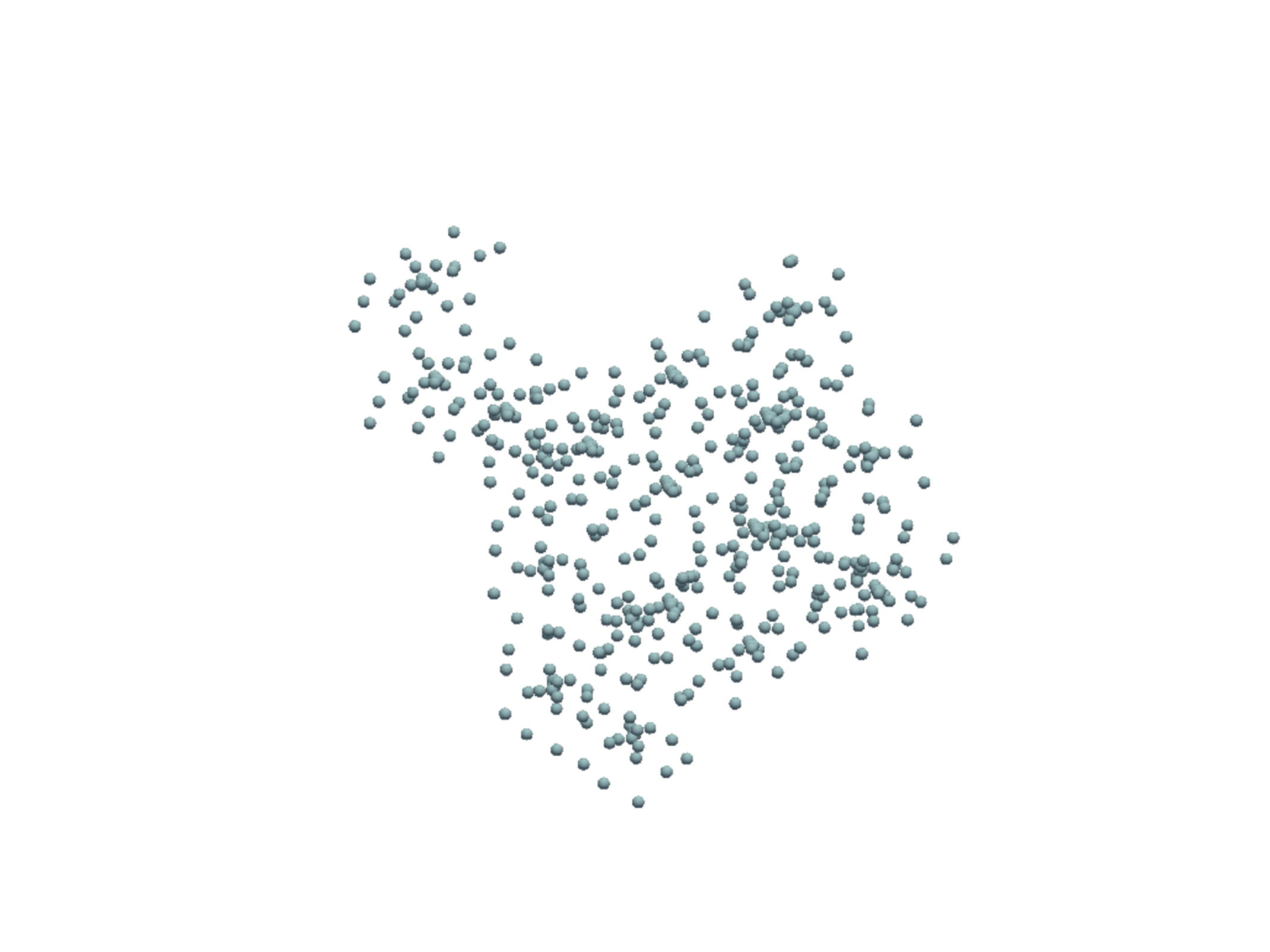} 
        \caption*{bpp: 0.26 - PSNR: 14.65}
    \end{subfigure}
    \hfill
    \begin{subfigure}{0.22\textwidth}
        \centering
        \includegraphics[width=.82\linewidth]{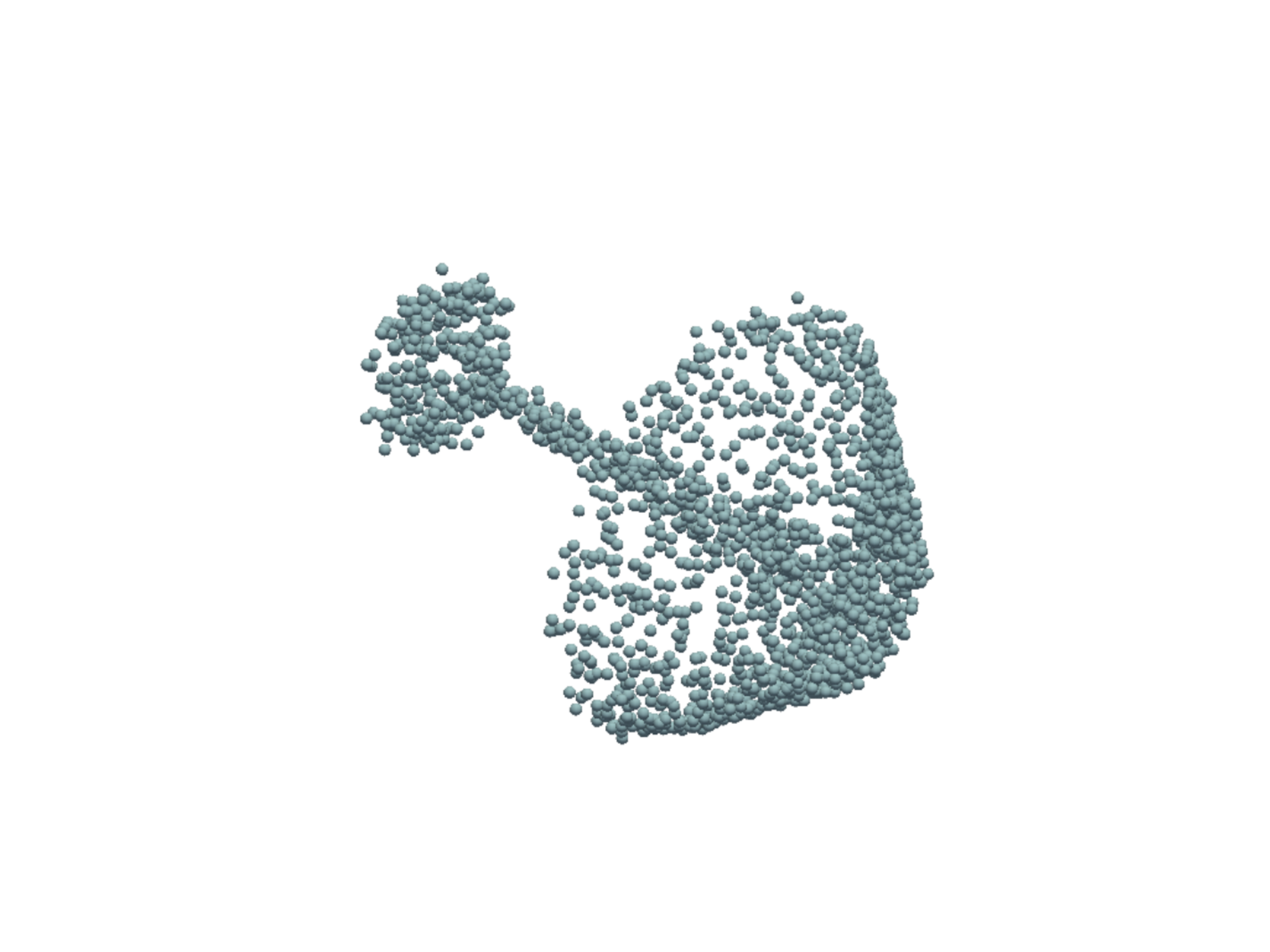} 
        \caption*{bpp: 0.25 - PSNR: 30.53}
    \end{subfigure}
    
    \vspace{0.5cm}
    
    \begin{subfigure}{0.22\textwidth}
        \centering
        \includegraphics[width=.82\linewidth]{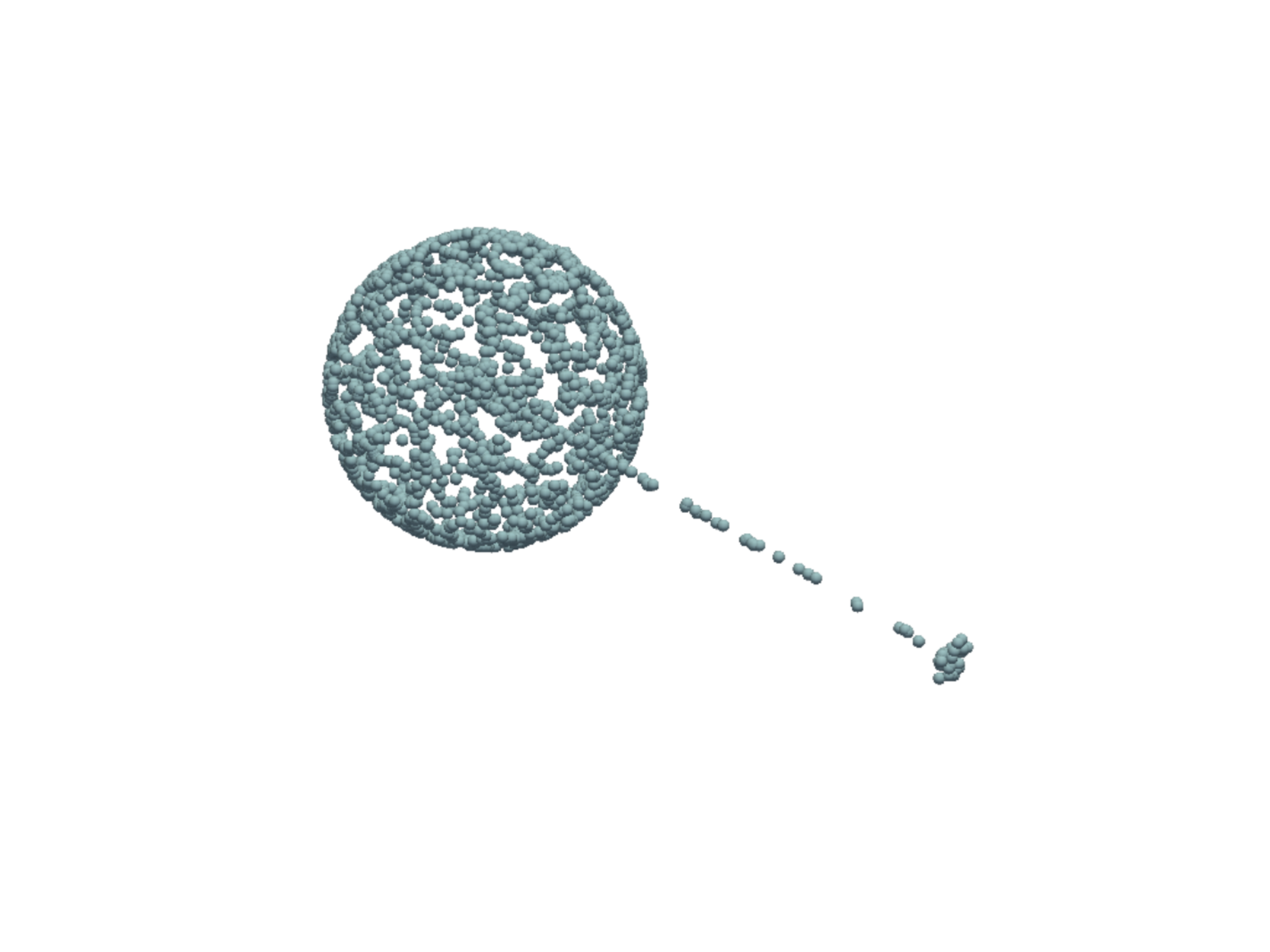} 
        \caption*{--}
    \end{subfigure}
    \hfill
    \begin{subfigure}{0.22\textwidth}
        \centering
        \includegraphics[width=.82\linewidth]{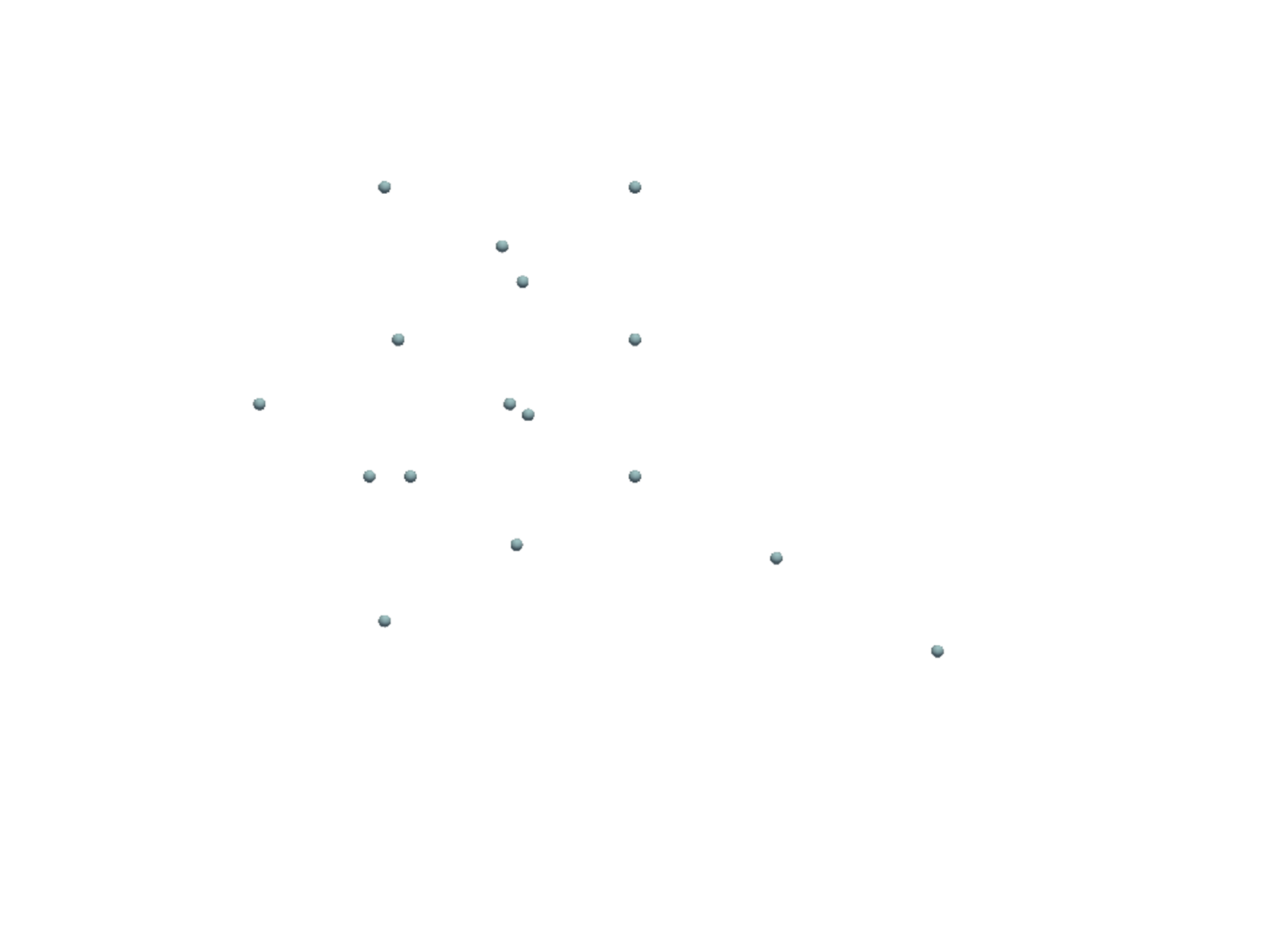} 
        \caption*{bpp: 0.0625 - PSNR: 13.30}
    \end{subfigure}
    \hfill
    \begin{subfigure}{0.22\textwidth}
        \centering
        \includegraphics[width=.82\linewidth]{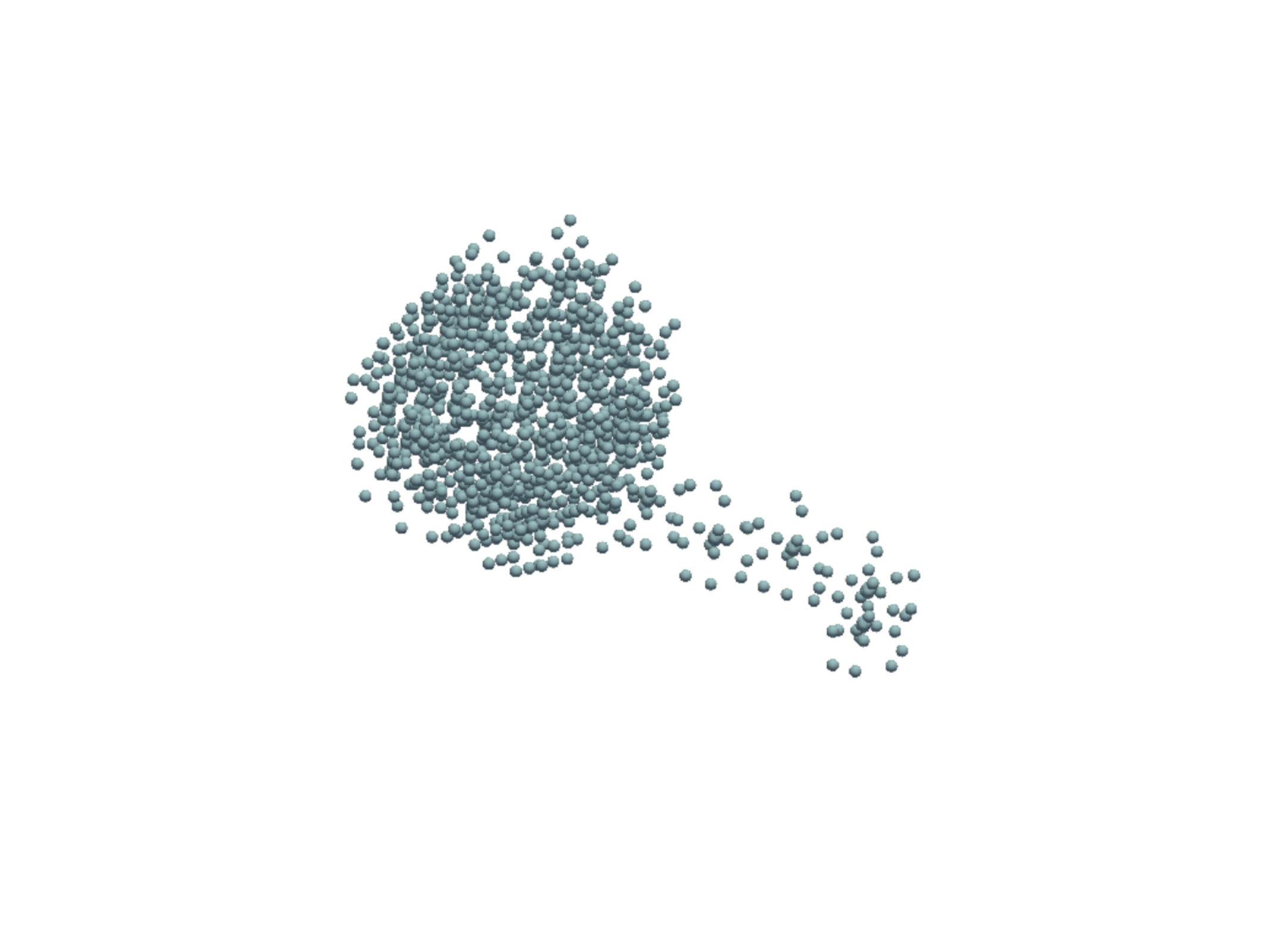} 
        \caption*{bpp: 0.52 - PSNR: 20.89}
    \end{subfigure}
    \hfill
    \begin{subfigure}{0.22\textwidth}
        \centering
        \includegraphics[width=.82\linewidth]{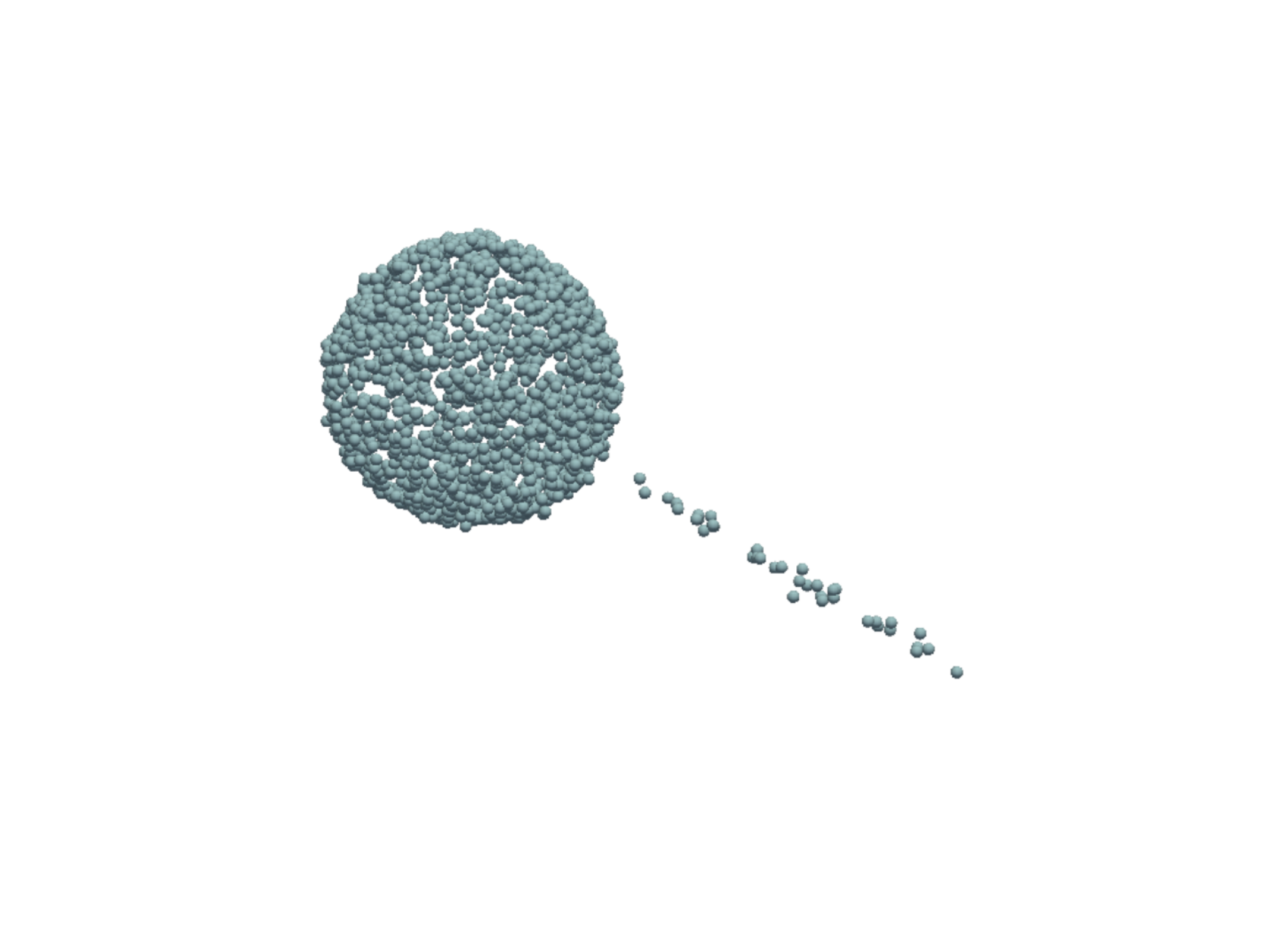} 
        \caption*{bpp: 0.0625 - PSNR: 33.80}
    \end{subfigure}
    
    \vspace{0.5cm}
    
    \begin{subfigure}{0.22\textwidth}
        \centering
        \includegraphics[width=.82\linewidth]{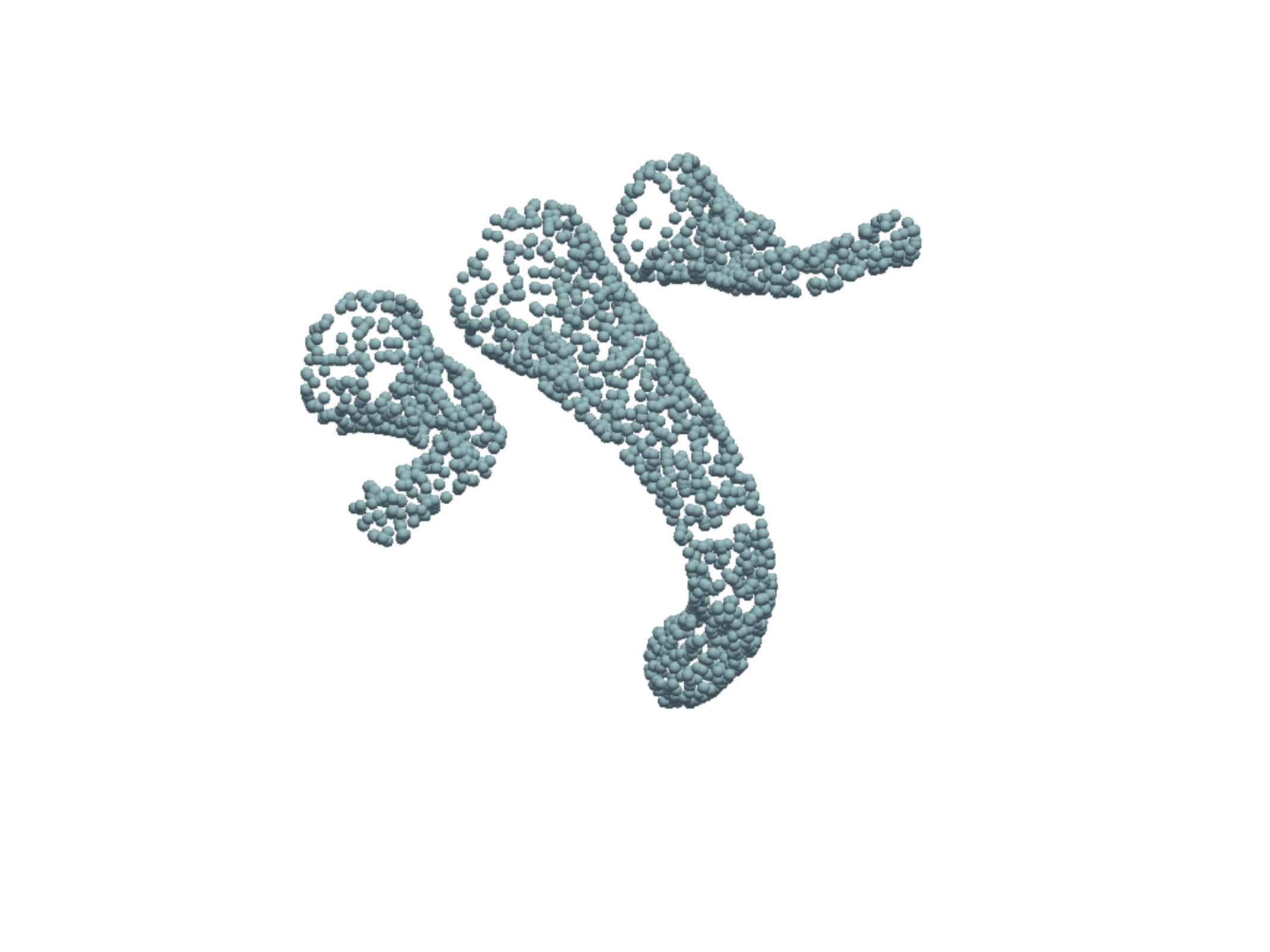} 
        \caption*{--}
    \end{subfigure}
    \hfill
    \begin{subfigure}{0.22\textwidth}
        \centering
        \includegraphics[width=.82\linewidth]{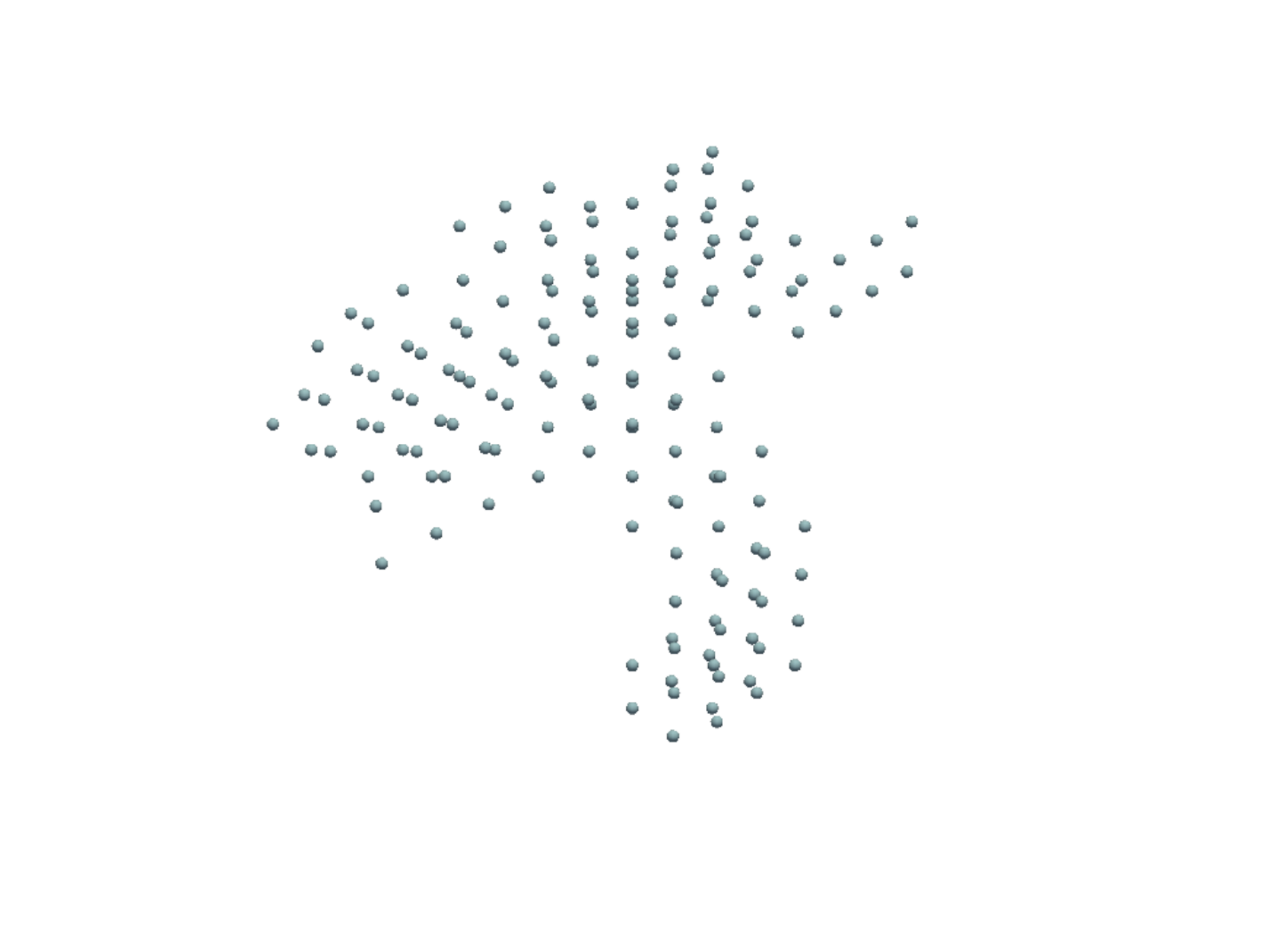} 
        \caption*{bpp: 0.23 - PSNR: 23.90}
    \end{subfigure}
    \hfill
    \begin{subfigure}{0.22\textwidth}
        \centering
        \includegraphics[width=.82\linewidth]{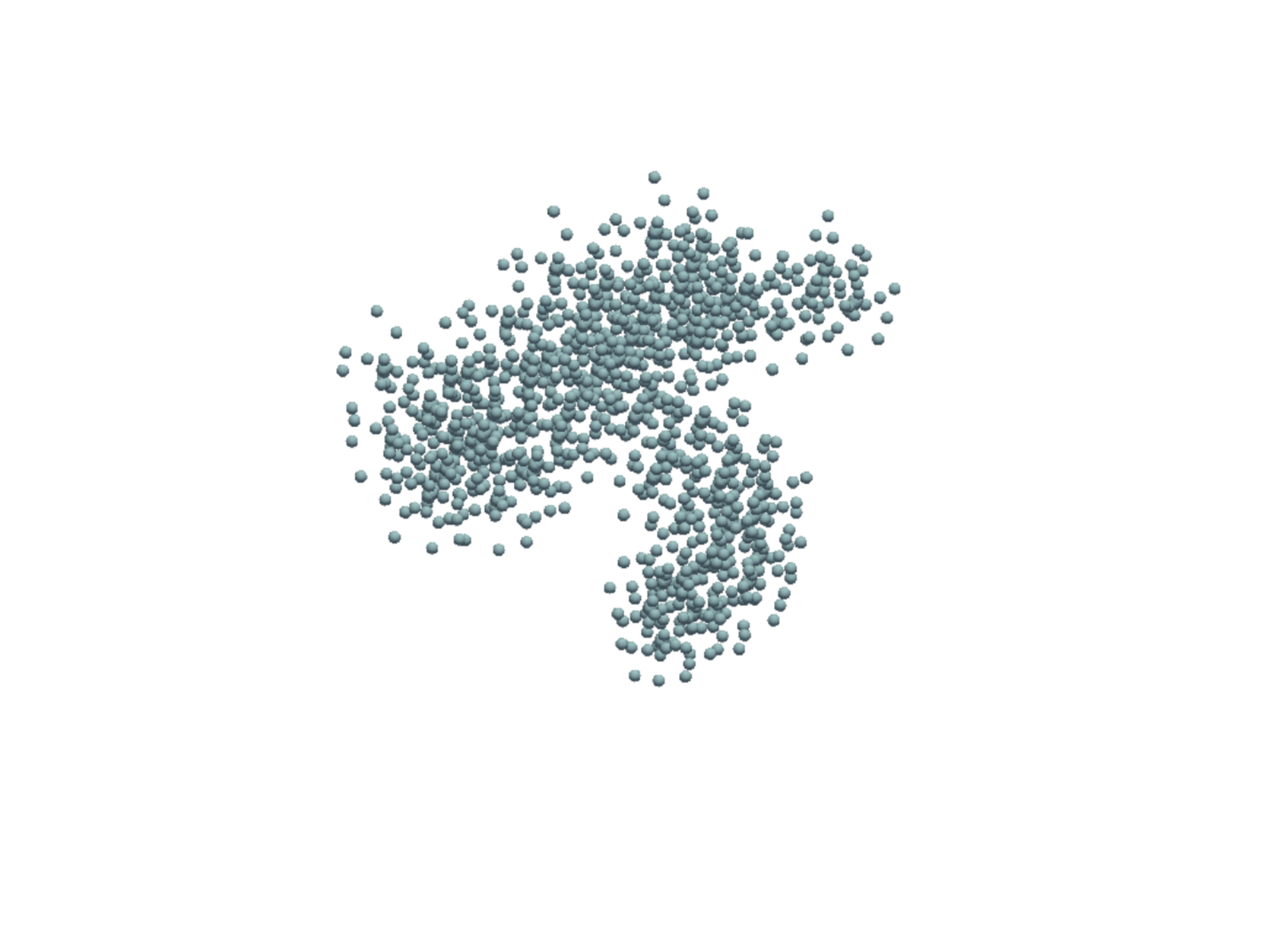} 
        \caption*{bpp: 0.52 - PSNR: 21.03}
    \end{subfigure}
    \hfill
    \begin{subfigure}{0.22\textwidth}
        \centering
        \includegraphics[width=.82\linewidth]{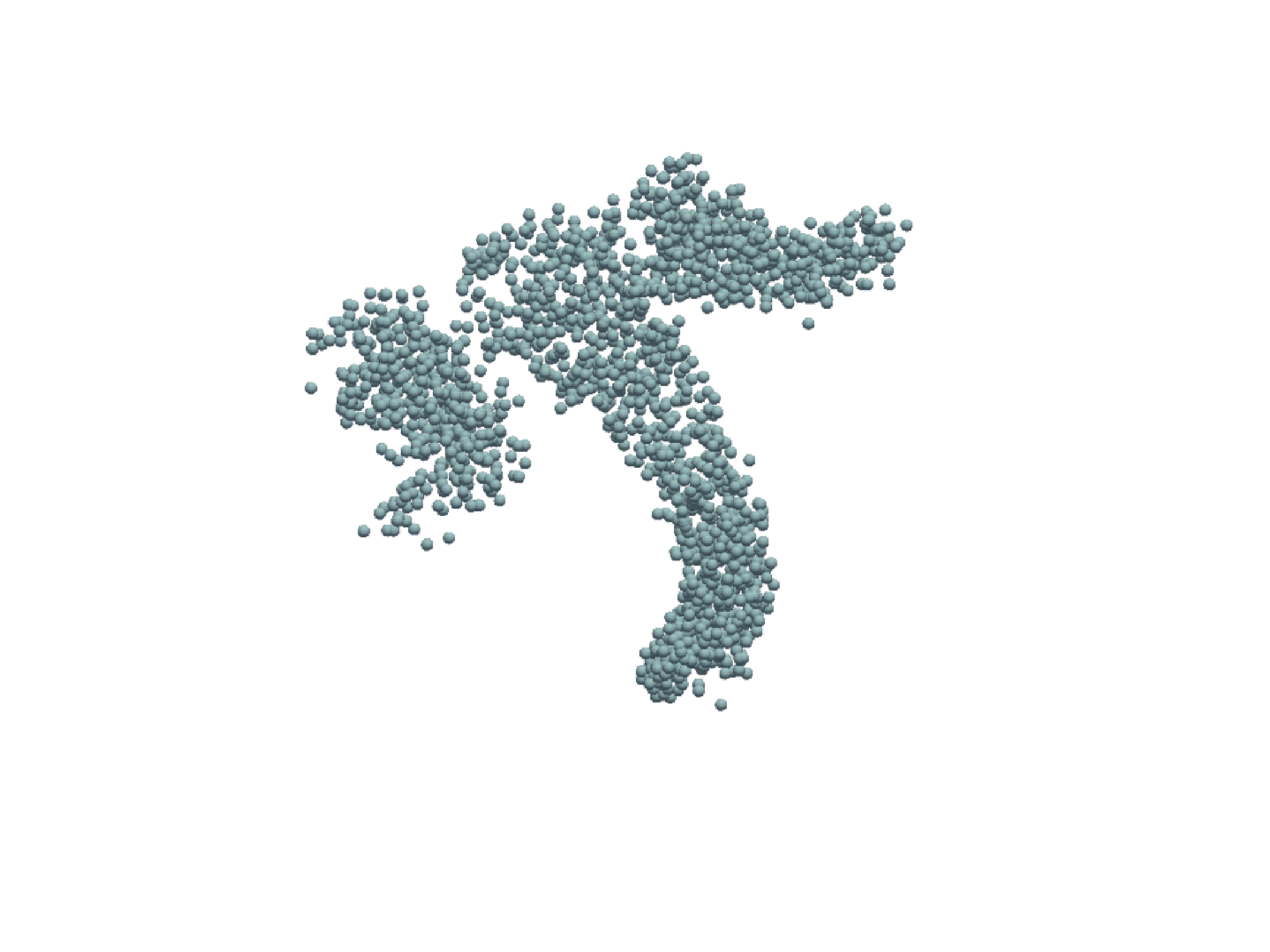} 
        \caption*{bpp: 0.25 - PSNR: 26.76}
    \end{subfigure}
    
    \caption{Qualitative results on ModelNet (first two rows) and ShapeNet (last two rows). We compare the reconstruction results of different compression methods providing the bpp and the PSNR values for each analyzed sample.}
    \label{fig:qualitative}
    \vspace{-1em}
    
\end{figure*}

%% file: sec/5_conclusion.tex
\section{Conclusions \& Limitations}
\label{sec:conclusions}
In this work we proposed DDPM-PCC, a simple yet effective model to compress the point cloud geometry at very low bit rates. Our compression model consists of a PointNet~\cite{pointnet} encoder used to generate a latent representation. This vector is then quantized with a learnable quantizer strategy and used to condition the diffusion reverse process to generate the input point cloud. The compression rate is then controlled during the quantization, dividing the latent code in a certain number of chucks and quantizing each of these sub-vectors independently. This allows us to compress the input at very low bit rates.
Despite its simplicity, the generation quality is upper-bounded by the performance of the generative model. A possible solution to increase the fidelity of the reconstruction at higher bit rates could be to provide side information regarding the geometry of the input point cloud to guide the generation.
\\[1em]
\noindent
\textbf{Acknowledgements. }  
This research was partially funded by Hi!PARIS Center on Data Analytics and Artificial Intelligence. This project was provided with computer and storage resources by GENCI at IDRIS thanks to the grant 2024-AD011015338 on the supercomputer Jean Zay.